\documentclass[runningheads]{llncs}
\usepackage{graphicx}
\usepackage{comment}
\usepackage{amsmath,amssymb} 
\usepackage{color}
\usepackage{arydshln}
\usepackage{booktabs}
\usepackage{url}
\usepackage{microtype}
\usepackage{refcount}

\makeatletter
\g@addto@macro{\UrlBreaks}{\UrlOrds}
\makeatother

\definecolor{mygray}{gray}{0.6}
\newcommand{\gray}[1]{\textcolor{mygray}{#1}}

\begin{document}
\pagestyle{headings}
\mainmatter
\def\ECCVSubNumber{5308}  

\title{What makes fake images detectable? Understanding properties that generalize} 

\titlerunning{What makes fake images detectable?}
%
\author{Lucy Chai\inst{1}\and
David Bau\inst{1} \and
Ser-Nam Lim\inst{2} \and Phillip Isola\inst{1}}
\authorrunning{Chai et al.}
%
\institute{MIT CSAIL, Cambridge MA, 02139\\
\email{\{lrchai,davidbau,phillipi\}@csail.mit.edu}\\
\and \email{sernam@gmail.com}}
\maketitle

\begin{abstract}

The quality of image generation and manipulation is reaching impressive levels, making it increasingly difficult for a human to distinguish between what is real and what is fake. However, deep networks can still pick up on the subtle artifacts in these doctored images. We seek to understand what properties of fake images make them detectable and identify what generalizes across different model architectures, datasets, and variations in training. We use a patch-based classifier with limited receptive fields to visualize which regions of fake images are more easily detectable. We further show a technique to exaggerate these detectable properties and demonstrate that, even when the image generator is adversarially finetuned against a fake image classifier, it is still imperfect and leaves detectable artifacts in certain image patches. Code is available at \url{https://chail.github.io/patch-forensics/}.

\end{abstract}

\section{Introduction}

State-of-the-art image synthesis algorithms are constantly evolving, creating a challenge for fake image detection methods to match the pace of content creation. It is straightforward to train a deep network to classify real and fake images, but of particular interest is the ability of fake image detectors to generalize to unseen fake images. What artifacts do these fake image detectors look at, and which properties can allow a detector released today to work on novel fake images? 

Generalization is highly desired in machine learning, with the hope that models work not only on training data, but also on related held-out examples as well. For tasks like object detection and classification, this has been accomplished with successively deeper and deeper networks that incorporate the context of the entire image to learn about global semantics and object characteristics. On the other hand, to learn image manipulation artifacts that are shared across various image generation pipelines, global content is not the only signal that matters. In fact, two identical generators trained on the same training data, differing only in the random initialization seed, can create differences in content detectable by a deep network classifier~\cite{yu2019attributing}. Instead of these differences, we seek to identify what image generators have \textit{in common}, so that training on examples generated from one model can help us identify fake images from another model. 

Across different facial image generators, we hypothesize that global errors can differ but local errors may transfer: the global facial structure can vary among different generators and datasets, but local patches of a generated face are more stereotyped and may share redundant artifacts. Therefore, these local errors can be captured by a classifier focusing on textures~\cite{geirhos2018imagenet} in small patches. We investigate a fully convolutional approach to training classifiers, allowing us to limit the receptive field of the model to focus on image patches. Furthermore, these patch-based predictions offer us a natural way to visualize patterns that are indicative of a real or fake image.

Using a suite of synthetic face datasets that span fully generative models~\cite{karras2017progressive,karras2019style,kingma2018glow,richardson2018gans} and facial manipulation methods~\cite{rossler2019faceforensics++}, we find that more complex patches, such as hair, are detectable across various synthetic image sources when training on images from a single source. In one of our early experiments, however, we observed that we could obtain misleadingly high generalization simply due to subtle differences in image preprocessing -- therefore, we introduce careful preprocessing to avoid simply learning differences in image formatting.

With a fixed classifier, an attacker can simply modify the generator to create adversarial examples of fake images, forcing them to become misclassified. Accordingly, we finetune a GAN to create these adversarial examples. We then show that a newly trained classifier can still detect images from this modified GAN, and we investigate properties of these detected patches. Our results here suggest that creating a coherent fake image without any traces of local artifacts is difficult: the modified generator is still unable to faithfully model certain regions of a fake image in a way that is indistinguishable from real ones.

Detecting fake images is a constant adversarial game with a number of ethical considerations. As of today, no method is completely bulletproof. Better generators, out-of-distribution images, or adversarial attacks~\cite{gragnaniello2018analysis,carlini2020evading} can defeat a fake-image detector, and our approach remains vulnerable to many of these same shortcomings. Furthermore, we train on widely used standard face datasets, but these are still images of real individuals. To protect the privacy of people in the dataset, we blur all real faces and manipulated real faces used in our figures. Our contributions are summarized as follows:
\begin{itemize}
    \item To avoid learning image formatting artifacts, we preprocess our images to reduce formatting differences between real and fake images.
    \item  We use a fully-convolutional patch-based classifier to focus on local patches rather than global structure, and test on different model resolutions, initialization seeds, network architectures, and image datasets. On facial images, we find that patch-based classifiers often perform better on out-of-domain synthetic images than full-image classifiers.
    \item We categorize the patches that are most indicative of real or fake images across various test datasets.
    \item To visualize detectable properties of fake images, we manipulate the generated images to exaggerate characteristic attributes of fake images.
    \item Finetuned generators are able to overcome a fake-image detector, but a subsequent classifier shows that detectable mistakes still occur in certain image patches.
\end{itemize}

\section{Related Work}

\subsubsection{Image manipulation.}

Verifying image authenticity is not just a modern problem --  historical instances of photo manipulation include a well-known portrait of Abraham Lincoln\footnote{\url{https://www.bbc.com/future/article/20170629-the-hidden-signs-that-can-reveal-if-a-photo-is-fake}} and instances of image censorship in the former Soviet Union\footnote{\url{https://en.wikipedia.org/wiki/Censorship_of_images_in_the_Soviet_Union}}.
However, recent developments in graphics and deep learning make creating forged images easier than ever. One of these manipulation techniques is image splicing, which combines multiple images to form a composite~\cite{hays2007scene}. This approach is directly relevant to face swapping, where a source face is swapped and blended onto a target background to make a person appear in a falsified setting. The deep learning analogue of face swapping, Deepfakes~\cite{deepfakesgithub}, has been the focus of much recent media attention. In parallel, improvements in generative adversarial networks (GANs) form another threat, as they are now able to create shockingly realistic images of faces simply from random Gaussian noise~\cite{karras2017progressive,karras2019style}.

\vspace{-0.1in}
\subsubsection{Automating detection of manipulated images.}
Given the ease in creating manipulated images nowadays and the potential to use them for malicious purposes, a number of efforts have focused on automating detection of manipulated images. A possible solution involves checking for consistency throughout the image -- examples include predicting metadata~\cite{huh2018fighting} or other low-level artifacts~\cite{popescu2004exposing,popescu2005resampling,popescu2005exposing}, learning similar embeddings for nearby patches~\cite{zhou2017two}, or learning similarity graphs from image patches~\cite{mayer2019exposing}. Other works have focused on training classifiers for the detection task, using a deep network either directly on RGB images~\cite{bayar2016deep,afchar2018mesonet,rossler2019faceforensics++} or alternative image representations~\cite{cozzolino2017recasting,mo2018fake}. ~\cite{rahmouni2017distinguishing} uses a combination of both: a CNN to extract features over image patches and a separate classifier for prediction. Here we also take a patch-wise approach, and we use these patches to visualize the network decisions.

\vspace{-0.1in}
\subsubsection{Can detectors generalize?}
The class of potential manipulations is so large that it is infeasible to cover all possible cases. Can a detector learn to distinguish real and fake images from one source and transfer that knowledge to a different source? Preprocessing is one way to encourage generalization, such as using spectral features~\cite{zhang2019detecting} or adding blur and random noise~\cite{xuan2019generalization}.~\cite{wang2019cnn} generalizes across a wide variety of datasets simply by adding various levels of augmentation. Specialized architectures also help generalization: for example~\cite{cozzolino2018forensictransfer} uses an autoencoder with a bottleneck that encourages different embeddings for real and fake images. A challenge with generalization is that the classifiers are not explicitly trained on the domain they are tested on. ~\cite{li2018exposing} demonstrates that it is possible to simulate the domain of manipulated images; by applying warping to source images, they can detect deepfake images without using manipulated images in classifier training.~\cite{li2020face} further studies generalization across different facial manipulation techniques also using a simulated domain of blended real images. However, a remaining question is what features do these models rely on to transfer knowledge among different domains, which we seek to investigate here. 

\subsubsection{Classification with local receptive fields.}

We use patch-based classification to visualize properties that generalize. Small receptive fields encourage the classifier to focus on local artifacts rather than global semantics, which is also an approach taken in GAN discriminators to encourage synthesis of realistic detailed textures~\cite{isola2017image}. A related concept is the Markovian generative adversarial network for texture synthesis~\cite{li2016precomputed}; the limited receptive field makes the assumption that only pixels within a certain radius affect the output, and the pixels outside that radius are independent from the output.~\cite{long2015fully} demonstrate a method for converting deep neural classifiers to fully convolutional networks and use patch-wise training, allowing the model to scale efficiently to arbitrarily-sized inputs, used for the task of semantic segmentation.

\section{Using Patches for Image Forensics}

Rather than training a network to predict a global ``real'' or ``fake'' decisions for an image, we use shallow networks with limited receptive fields that focus on small patches of the image. This approach allows us to localize regions of the image that are detected to be manipulated and ensemble the patch-wise decisions to obtain the overall prediction.

\begin{figure}[t]
\centering
\includegraphics[width=0.9\textwidth]{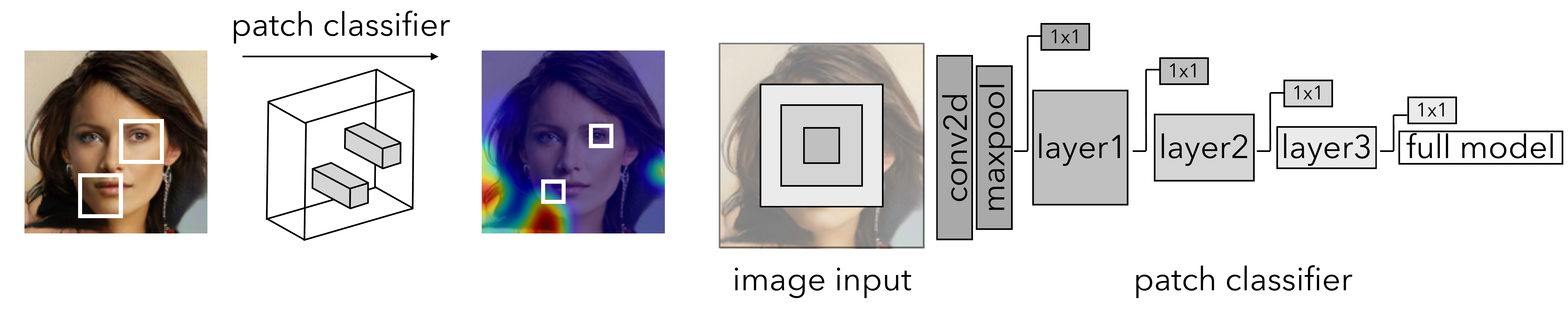} 
\caption{We use a classifier with small receptive fields to obtain a heatmap over the patch-wise output. To obtain this patch classifier, we truncate various deep learning models after an initial sequence of layers.}
\label{fig:schematic}
\vspace{-0.1in}
\end{figure}

\subsection{Models for patch-based classification.}

Modern deep learning architectures typically consist of a series of modular blocks. By truncating the models after an intermediate block, we can obtain model predictions based on a local region of the image, where truncating earlier in the layer sequence results in a smaller receptive field, while truncating after more layers results in a larger receptive field. We then add a 1x1 convolution layer after this truncated backbone to convert the feature representation into a binary real-or-fake prediction. We experiment with Resnet and Xception as our model backbones -- generally we observe that Xception blocks perform better than Resnet blocks, however we also report results of the top performing Resnet block. We provide additional details on the model architecture and receptive field calculations in Supplementary Material Sec.~\ref{apx:architecture}.

The truncation operation reduces the size of the model's receptive field, and yields a prediction for a receptive-field-sized patch of the input, rather than the entire image at once. This forces the models to learn local properties that distinguish between real and fake images, where the same model weights are applied in a sliding fashion over the entire image, and each output prediction is only a function of a small localized patch of the image. We apply a cross entropy loss to each patch; i.e. every real patch should be considered real, and every fake image patch should be considered fake:
\begin{equation}
    \mathcal{L}(\mathbf{x}) = \frac{1}{|P|}\sum_{i,j}\sum_t t\log f^t(x_{i,j})
\end{equation}
where $f$ is the model output after a softmax operation to normalize the logits, $t$ indexes over the real and fake output for binary classification, $(i,j)$ indexes over the receptive field patches, and $|P|$ is the total number of patches per image. We train these models with the Adam optimizer with default learning rate, and terminate training when validation accuracy does not improve for a predetermined number of epochs.

By learning to classify patches, we increase the ratio of data points to model parameters: each patch of the image is treated independently, and the truncated models are smaller. The final classification output is an ensemble of the individual patch decisions rather than a single output probability. To aggregate patches at inference time, we take a simple average after applying a softmax operation to the patch-wise predictions: 
\begin{equation}
    t^* = \arg\max_t  \left(\frac{1}{|P|}\sum_{i,j} f^t(x_{i,j})\right)
\end{equation}
The averaging approach can be applied in both cases where the image is wholly generated, or when only part of the image is manipulated. For example, when a generated face is spliced onto a real background, the background patches may not be predicted as fake; in this case, because the same background is present in both real and fake examples, the model remains uncertain in these locations.

\begin{figure}[t]
\centering
\includegraphics[width=0.9\textwidth]{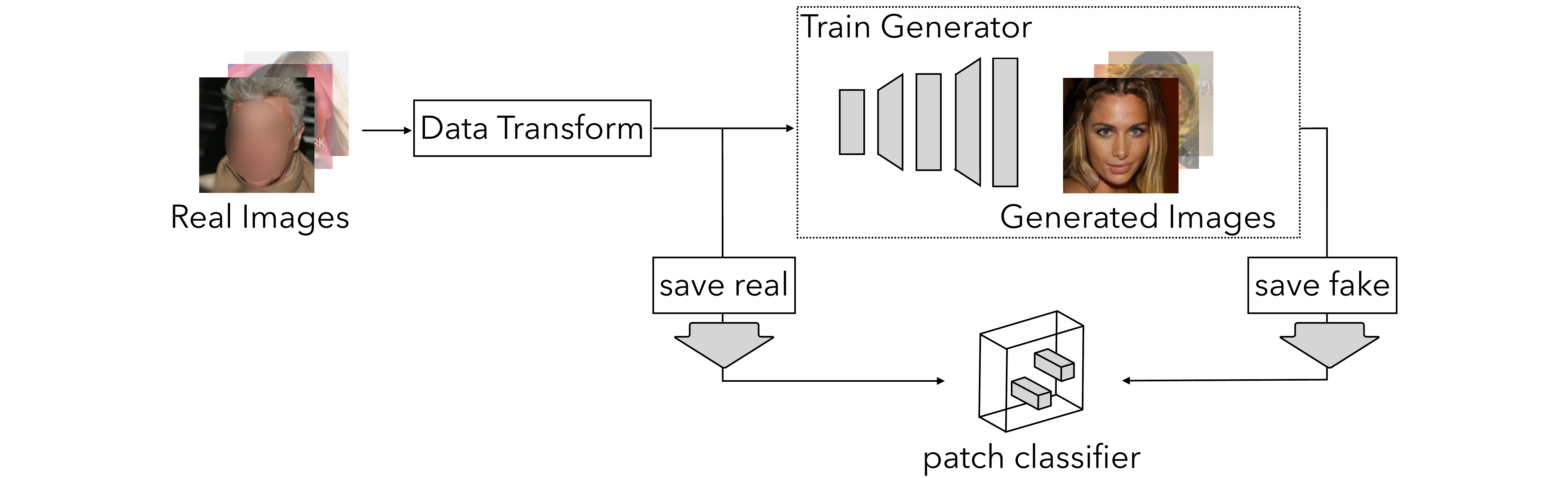} 
\caption{To minimize the effect of image preprocessing artifacts - in which real and fake images undergo different preprocessing operations -- we pass real images through the same data transform as used to train the generator. We then save the real and fake images using identical pipelines.}
\label{fig:preprocessing}
\vspace{-0.1in}
\end{figure}

\subsection{Dataset preparation} 

\subsubsection{Image preprocessing.} A challenge with fully generative images, such as those created by GANs, is that fake images can be saved with arbitrary codecs, e.g., we decide whether we want to save the image in JPG or PNG format. However, the set of real images is saved with a fixed codec when the original dataset is created. When training a classifier on real and fake images with subtly different preprocessing pipelines, the classifier can simply learn to detect the differences in preprocessing. If the test images also have this inconsistency, we would appear to obtain high accuracy on the test set, even though the classifier is really only detecting formatting artifacts. One way to mitigate this disparity is to apply data augmentation to reduce the effect of these differences~\cite{wang2019cnn,xuan2019generalization}.

We preprocess the images to make our real and fake dataset as similar as possible, in an effort to isolate fake image artifacts and minimize the possibility of learning differences in preprocessing. We create the ``real'' dataset by passing the real images through the generator's data loading pipeline (e.g. resizing) and saving the real images \textit{after} this step in lossless PNG format (Fig.~\ref{fig:preprocessing}). We save the fake images in PNG format from the generator output, so the remaining differences between real and fake images are due to artifacts of the generator. We then resize all images to the same size using Lanczos interpolation before saving to file. Additional details are provided in Supplementary Material Sec.~\ref{apx:preprocessing}.

We take these precautions because any minor difference in preprocessing is easily learnt by the fake-image classifier and leads to an illusion of increased generalization capacity (for example, differences in the image codec leads to perfect average precision across various test datasets; see Supplementary Material Sec.~\ref{apx:preprocessing}). This approach allows us to focus on the inherent differences between real images and generated ones to minimize any potential confounders due to preprocessing. In the remainder of this section, we briefly detail the image generation and manipulation methods that we investigate in our experiments.

\vspace{-0.1in}
\subsubsection{Fully Generative Models.} The first class of models we consider are fully generative models which map a random sample from a known distribution (e.g. a multivariate Gaussian) to an image. \textit{Progressive GAN} (PGAN)~\cite{karras2017progressive} is one recent example, which uses a progressive training schedule to increase the output resolution of images during training. We use the publicly available PGAN model trained on the CelebA-HQ face dataset. We also train several other PGANs to various smaller resolutions and on the more diverse FFHQ face dataset. \textit{StyleGAN} (SGAN)~\cite{karras2019style} introduces an alternative generator architecture which incorporates the latent code into intermediate layers of the generator, resulting in unsupervised disentanglement of high-level attributes, e.g., hair and skin tone.
We use the public versions of StyleGAN on the CelebA-HQ and FFHQ datasets, and StyleGAN2 \cite{karras2019analyzing} on the FFHQ dataset. In additional to PGAN and SGAN, we also consider the \textit{Glow} generator~\cite{kingma2018glow}, a flow-based model using modified 1x1 invertible convolutions that optimizes directly for log-likelihood rather than adversarial loss. We use the public Glow generator trained on CelebA-HQ faces. Finally, we also include a face generator based on a \textit{Gaussian Mixture Model} (GMM) rather than convolutional layers~\cite{richardson2018gans}; the GMM uses low-rank plus diagonal Gaussians to efficiently model covariance in high-dimensional outputs such as images. We train the GMM model on the CelebA~\cite{liu2015faceattributes} dataset using default parameters.

\subsubsection{Facial Manipulation Models.} We use the FaceForensics++ dataset~\cite{rossler2019faceforensics++}, which includes methods for identity manipulation and expression transfer. Identity manipulation approaches, such as \textit{FaceSwap}, paste a source face onto a target background; specifically, FaceSwap fits detected facial landmarks to 3D model and then projects the face onto the target scene. The deep learning analogue to FaceSwap is the \textit{Deepfake} technique, which uses a pair of autoencoders with a shared encoder to swap the source and target faces. On the other hand, expression transfer maps the expression of a source actor onto the face of a target. \textit{Face2Face} achieves this by tracking expression parameters of the face in a source video and applying them to a target sequence. \textit{Neural Textures} uses deep networks to learn a texture map and a neural renderer to modify the expression of the target face. 
\subsection{Baseline Models.}

We train and evaluate full MesoInception4~\cite{afchar2018mesonet}, Resnet~\cite{he2016deep}, and Xception~\cite{chollet2017xception} models on the same datasets that we use to train the truncated classifiers. Following~\cite{afchar2018mesonet}, we train MesoInception4 using squared error loss. For the Resnet model and the Xception model, also used in~\cite{rossler2019faceforensics++}, we train with standard two-class cross entropy loss. We train these models from scratch as they are not initially trained for this classification task. Finally, we also compare to a model trained to detect CNN artifacts via blurring and compression augmentations~\cite{wang2019cnn}. For this model, we finetune at a learning rate of 1e-6 using similar augmentation parameters as the original paper to improve its performance specifically on face datasets. We use the same stopping criteria based on validation accuracy for all baseline models as we use for the truncated models.

\section{Experiments}

\subsection{Classification via patches.}

Nowadays with access to public source code, it becomes easy for anyone to train their own image generators with slight modifications. We conduct two experiments to test generalization across simple changes in (1) generator size and (2) the weight initialization seed. In addition to the public 1024px PGAN, we train PGANs for 512, 256, and 128px resolutions on the CelebA-HQ dataset and sample images from each generator. We then train a classifier using only images from the 128px generator. 

We test the classifier on generated images from the remaining resolutions, using average precision (AP) as a metric (Table~\ref{table:resolution}; left). Here, the full-model baselines tend to perform worse on the unseen test resolutions compared to the truncated models. However, adding blur and JPEG augmentations in \cite{wang2019cnn} helps to overcome the full-model limitations, likely hiding the resizing artifacts. Of the truncated models, the AP tends to decrease on the unseen test images as the receptive field increases, although there is a slight decline when the receptive field is too small with the Xception Block 1 model. On average across all resolutions, the Xception Block 2 model obtains highest AP.

\setlength{\tabcolsep}{3pt}
\begin{table}[t]
\begin{center}
\caption{Average precision across PGANs trained to different resolutions or with different random initialization seeds. The classifier is trained on a fake images from a 128px GAN and real images at 128px resolution. AP on the test set corresponding to training images is colored in gray.}
\label{table:resolution}
\resizebox{0.9\linewidth}{!}{
\begin{tabular}{l cccc  cccc}
\toprule
\multicolumn{1}{c}{} &\multicolumn{4}{c}{\textbf{Resolution}}& \multicolumn{4}{c}{\textbf{Model Seed}} \\
\cmidrule(lr){2-5} \cmidrule(lr){6-9} 
\textbf{Model Depth} & \textbf{128} & \textbf{256} & \textbf{512} & \textbf{1024} & \textbf{0} & \textbf{1} & \textbf{2} & \textbf{3} \\
\midrule
Resnet Layer 1 & \gray{100.0} & 99.99 & 99.60 & 96.95 & 
\gray{100.0} & 100.0 & 100.0 & 100.0\\
Xception Block 1 & \gray{100.0} & 100.0 & 99.87 & 98.53 &
\gray{100.0} & 100.0 & 100.0 & 100.0 \\
Xception Block 2 & \gray{100.0} & 100.0 & 100.0 & 99.98 &
\gray{100.0} & 100.0 & 100.0 & 100.0 \\
Xception Block 3 & \gray{100.0} & 100.0 & 100.0 & 99.92 &
\gray{100.0} & 100.0 & 100.0 & 100.0 \\
Xception Block 4 & \gray{100.0} & 100.0 & 99.92 & 99.34 &
\gray{100.0} & 100.0 & 100.0 & 100.0\\
Xception Block 5 & \gray{100.0} & 100.0 & 98.90 & 91.18 &
\gray{100.0} & 100.0 & 100.0 & 100.0 \\
\hdashline
\cite{afchar2018mesonet} MesoInception4 & \gray{100.0} & 99.59 & 98.15 & 87.00 & 
\gray{100.0} & 99.99 & 99.82 & 99.95 \\
\cite{he2016deep} Resnet-18 & \gray{99.99} & 96.85 & 91.75 & 80.17 & 
\gray{99.99} & 98.41 & 95.20 & 95.02 \\
\cite{chollet2017xception} Xception & \gray{100.0} & 99.94 & 99.84 & 97.28 & 
\gray{100.0} & 100.0 & 99.99 & 100.0 \\
\cite{wang2019cnn} CNN (p=0.1) & \gray{100.0} & 99.99 & 99.97 & 99.78 &
\gray{100.0} & 100.0 & 100.0 & 100.0 \\
\cite{wang2019cnn} CNN (p=0.5) & \gray{100.0} & 100.0 & 99.99 & 99.83 &
\gray{100.0} & 100.0 & 100.0 & 100.0 \\
\bottomrule
\end{tabular}
}
\end{center}
\vspace{-0.2in}
\end{table}
\setlength{\tabcolsep}{1.4pt}

Next, we train four PGANs to 128px resolution with different weight initialization seeds. We train the classifier using fake images drawn from one of the generators, and test on the remaining generators (Table~\ref{table:resolution}; right). Surprisingly, even when the only difference between generators is the random seed, the full Resnet-18 model makes errors when classifying fake images generated by the three other GANs. This suggests that fake images generated by different PGANs differ slightly between the different initialization seeds (as also noted in~\cite{yu2019attributing}). The MesoInception4 and Xception architectures are more robust to model seed, and so is blur/JPG augmentation. The truncated models with reduced receptive field are also robust to model seed differences. 

\setlength{\tabcolsep}{3pt}
\begin{table}[t]
\begin{center}
\caption{Average precision on different model architectures and an alternative dataset (FFHQ). The classifier is trained on 1024px PGAN random samples and reprojected PGAN images on the CelebA-HQ dataset. For the Glow model (*) we observe better performance when classifier training does not include reprojected images for the truncated models; additional results in Supplementary Material Sec. 2.4.  AP on the test set corresponding to training images is colored in gray.}
\label{table:model}
\vspace{-0.1in}
\resizebox{1.0\linewidth}{!}{
\begin{tabular}{lc ccc ccc}
\toprule
&\multicolumn{1}{c}{} & \multicolumn{3}{c}{\textbf{Architectures}} & \multicolumn{3}{c}{\textbf{FFHQ dataset}}\\
\cmidrule(lr){3-5}\cmidrule(lr){6-8}
\textbf{Model} & \textbf{PGAN} & \textbf{SGAN} & \textbf{Glow*} & \textbf{GMM} & \textbf{PGAN} & \textbf{SGAN} & \textbf{SGAN2} \\
\midrule
Resnet Layer 1 & \gray{100.0} & 97.22 & 72.80 & 80.69 & 99.81 & 72.91 & 71.81 \\
Xception Block 1 & \gray{100.0} & 98.68 & 95.48 & 76.21 & 99.68 & 81.35 & 77.40 \\
Xception Block 2 & \gray{100.0} & 99.99 & 67.49 & \textbf{91.38} & \textbf{100.0} & 90.12 & 90.85 \\ 
Xception Block 3 & \gray{100.0} & \textbf{100.0} & 74.98 & 80.96 & 100.0 & 92.91 & \textbf{91.45} \\
Xception Block 4 & \gray{100.0} & 99.99 & 66.79 & 42.82 & 100.0 & \textbf{95.85} & 90.62 \\
Xception Block 5 & \gray{100.0} & 100.0 & 60.44 & 48.92 & 100.0 & 93.09 & 89.08 \\
\hdashline
\cite{afchar2018mesonet} MesoInception4 & \gray{100.0} & 97.90 & 49.72	 & 45.98 & 98.71 & 80.57 & 71.27 \\
\cite{he2016deep} Resnet-18 &  \gray{100.0} & 64.80 & 47.06 & 54.69 & 79.20 & 51.15 & 52.37 \\
\cite{chollet2017xception} Xception & \gray{100.0} & 99.75 & 55.85 & 40.98 & 99.94 & 85.69 & 74.33 \\
\cite{wang2019cnn} CNN (p=0.1)  & \gray{100.0} & 98.41 & 90.46 & 50.65 & 99.95 & 90.48 & 85.27 \\
\cite{wang2019cnn} CNN (p=0.5)  & \gray{100.0} & 97.34 & \textbf{97.32} & 73.33 & 99.93 & 88.98 & 84.58 \\

\bottomrule
\end{tabular}
}
\end{center}
\vspace{-0.4in}
\end{table}
\setlength{\tabcolsep}{1.4pt}

We then test the ability of patch classifiers to generalize to different generator architectures (Table~\ref{table:model}; left). To create a training set of PGAN fake images, we combine two datasets -- random samples from the generator, as well as images obtained by reprojecting the real images into the GAN following \cite{bau2019seeing}. Intuitively, this reprojection step creates fake images generated by the GAN that are as close as possible to their corresponding real images, forcing the classifier to focus on the remaining differences (also see Supplementary Material Sec.~\ref{apx:reprojection},~\ref{apx:samples_inverse}). We then test the classifier on SGAN, Glow, and GMM face generators. 
We show additional results training on only PGAN fake samples, as well as only on reprojected images as the fake dataset in Supplementary Material Sec.~\ref{apx:samples_inverse} (on the Glow model, AP is substantially better when trained without the reprojected images). Generalizing to the SGAN architecture is easiest, due to the many similarities between the PGAN and SGAN generators. With the exception of the Glow generator, the truncated models obtain higher AP compared to the larger classifiers with a fraction of the number of parameters.


Lastly, we test the classifiers' ability to generalize to a different face dataset (Table~\ref{table:model}; right). Using the same classifiers trained on CelebA-HQ faces and PGAN samples and reprojections, we measure AP on real images from the FFHQ dataset and fake images from PGAN, SGAN, and SGAN2 trained on FFHQ faces. The truncated classifiers improve AP, particularly on the Style-based generators. The FFHQ dataset has greater diversity in faces than CelebA-HQ; however, small patches, such as hair, are likely similar between the two datasets. Using small receptive fields allows models to ignore global differences between images from different generators and datasets and focus on shared generator artifacts, perhaps explaining why truncated classifiers perform better than full models.


\begin{figure}[t]
\centering
\includegraphics[width=0.98\textwidth]{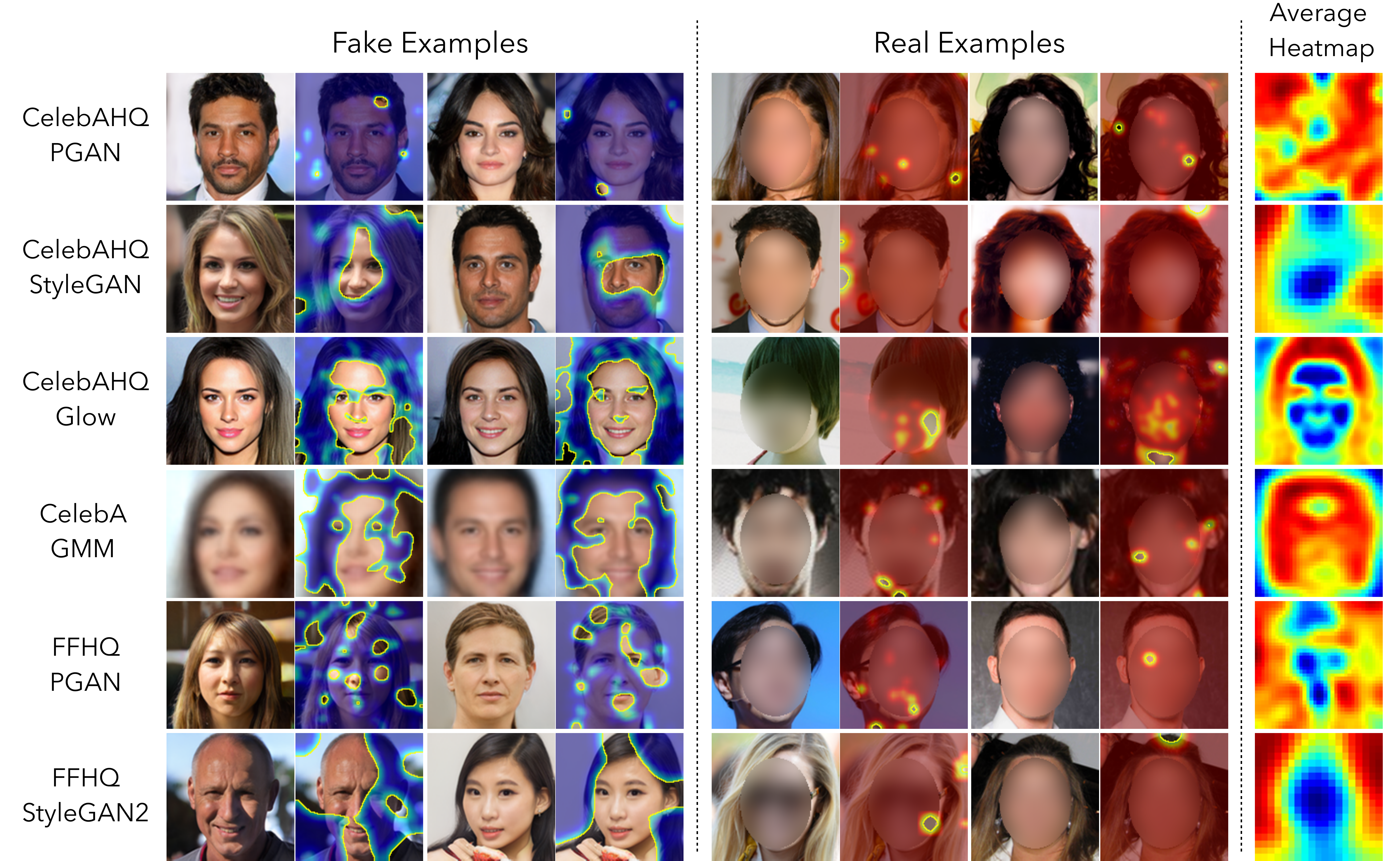} 
\caption{Heatmaps based on the patch-wise predictions on real and fake examples from each dataset and fake image generator. We normalize all heatmaps between 0 and 1 and show fake values in blue and real values in red. We also show the average heatmap over the 100 easiest and fake examples, where red is most indicative of the correct class.}
\label{fig:heatmaps}
\vspace{-0.2in}
\end{figure}

\begin{figure}[ht!]
\centering
\includegraphics[width=\textwidth]{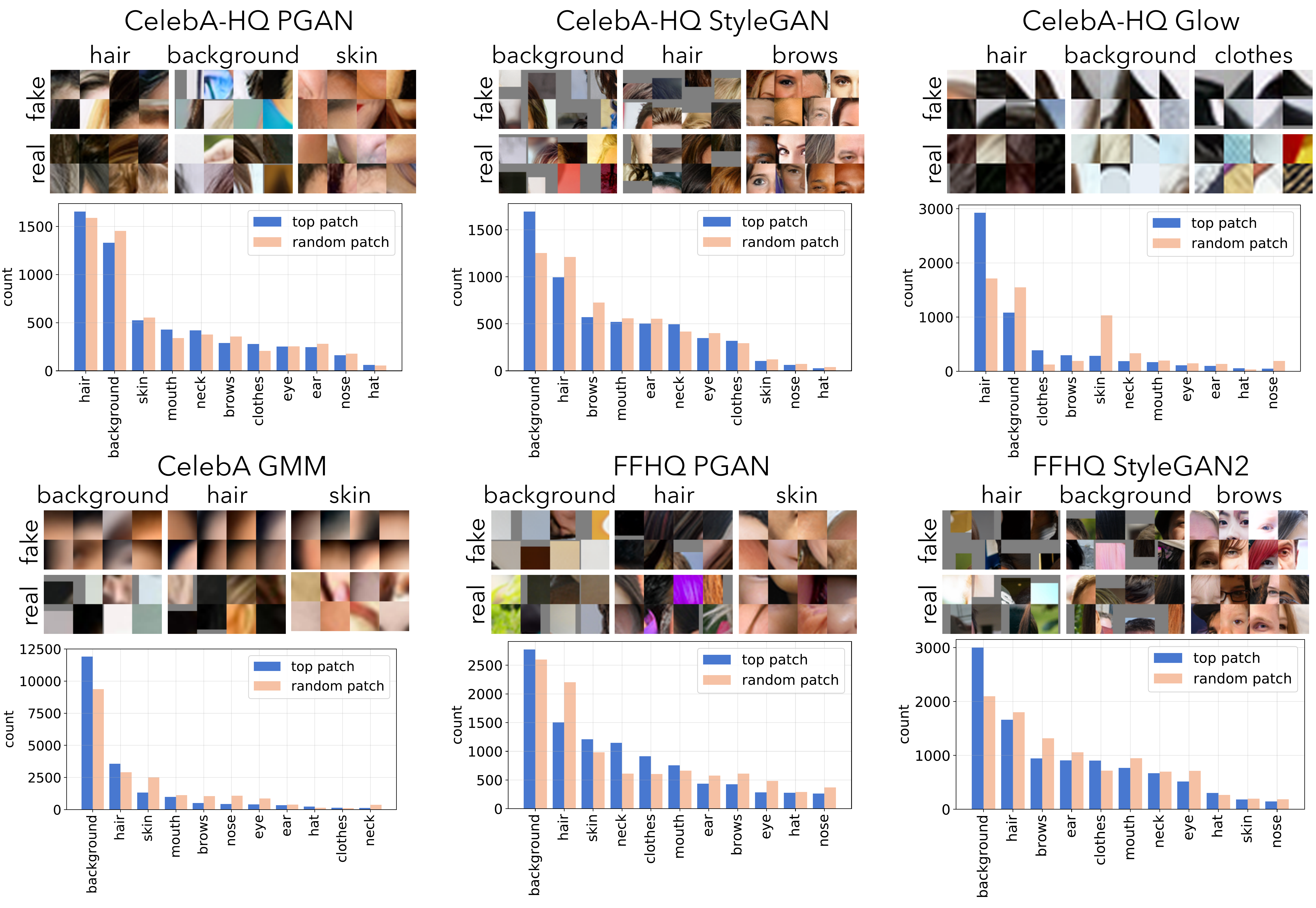} 
\vspace{-0.3in}
\caption{We take a pretrained segmentation network to assign the most predictive patch in real and fake images to a semantic cluster. We find that the fake-image classifier (which was only trained on the CelebA-HQ dataset with PGAN fake images) relies on patches such as hair, background, clothing, and mouths to make decisions.
}
\label{fig:histograms}
\vspace{-0.1in}
\end{figure}

\subsection{What properties of fake images generalize?}

What artifacts do classifiers learn that allow them to detect fake images generated from different models? Since the patch-based classifiers output real-or-fake predictions over sliding patches of a query image, we use these patch-wise predictions to draw heatmaps over the images and visualize what parts of an image are predicted as more real or more fake (Fig.~\ref{fig:heatmaps}). Using the classifiers trained on CelebA-HQ PGAN images, we show examples of the prediction heatmaps for the other face generators and on the FFHQ dataset, using the best performing patch model for each column in Table~\ref{table:model}.  We also show an averaged heatmap  over the 100 most real and most fake images, where the red areas indicate regions most indicative of the correct class (Fig.~\ref{fig:heatmaps}; right). The average heatmaps highlight predominately hair and background areas, indicating that these are the regions that patch-wise models rely on when classifying images from unseen test sources.

\begin{figure}[ht]
\centering
\includegraphics[width=\textwidth]{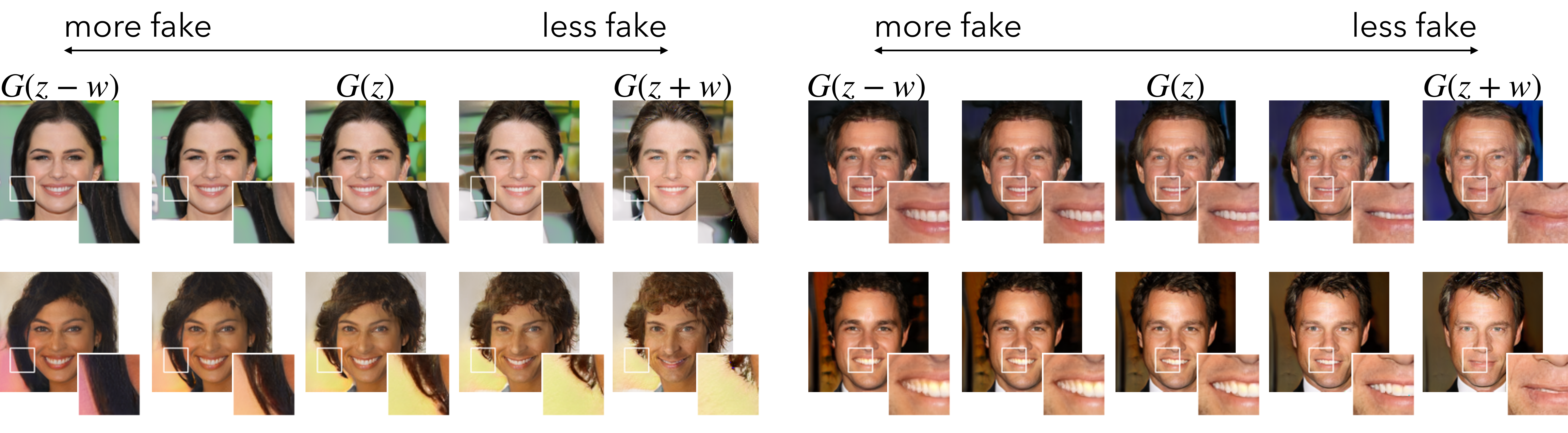} 
\caption{We shift the latent space of the PGAN generator to exaggerate the fake features of an image, which exaggerates the hair and smile of the fake images.}
\label{fig:caricatures}
\end{figure}

Next, we take a pretrained facial segmentation network to partition each image into semantic classes. For the most predictive patch in each image, we assign the patch to a cluster from the segmentation map, and plot the distribution of these semantic clusters (Fig.~\ref{fig:histograms}). We also sample a random patch in each image and assign it to a semantic cluster for comparison. Using the segmentation model, the predominant category of patches tends to be hair or background, with clothes, skin, or brows comprising the third-largest category. Qualitatively, many of the fake patches contain boundary edges such as those between hair and background or hair and skin, suggesting that creating a realistic boundary is difficult for image generators to imitate, whereas crisp boundaries naturally exist in real images.

To further understand what makes fake images look fake, we modify the latent space of the PGAN generator to exaggerate the features that the classifier detects (Fig.~\ref{fig:caricatures}). We parametrize a shift in latent space by a vector $w$, and optimize:
\begin{equation}
    w^* = \arg \min_w \mathbb{E}_z \left[ L_{\mathrm{fake}}(G(z-w)) + L_p(G(z),\;G(z-w)) \right]
\end{equation}
where $L_{\mathrm{fake}}$ refers to the classifier loss on fake images~\cite{goetschalckx2019ganalyze}, and $L_p$ is a perceptual loss regularizer~\cite{zhang2018unreasonable} to ensure that the modified image does not deviate too far from the original. Applying this vector to latent space samples accentuates hair and smiling with teeth, which are both complex textures and likely difficult for generators to recreate perfectly (Fig.~\ref{fig:caricatures}). By applying the shift in the opposite direction, $G(z+w)$, we see a reduction these textures, in effect minimizing the presence of textures that are more challenging for the generator to imitate.  

\begin{figure}[t]
\centering
\includegraphics[width=\textwidth]{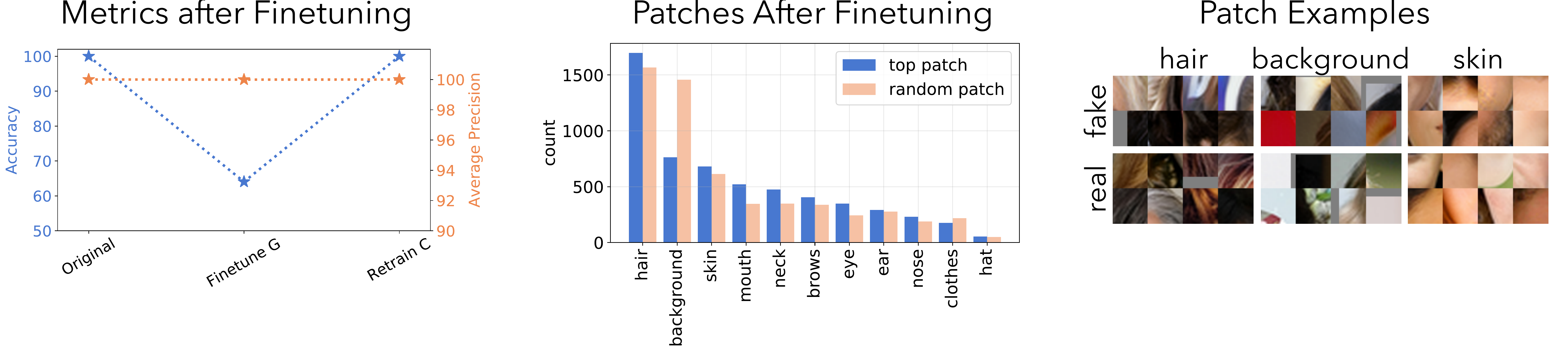} 
\vspace{-0.3in}
\caption{We finetune the PGAN generator to evade detection by the fakeness classifier. When we subsequently train a new classifier, we find that the finetuned generator still has detectable artifacts, but now predominantely less in background patches.}
\label{fig:finetune}
\vspace{-0.1in}
\end{figure}

\subsection{Finetuning the generator.} 

With access to gradients from the classifier, an easy adversarial attack is to modify the generator to evade detection by the classifier. Will this now make the previously identified fake patches undetectable? To investigate this, we finetune a PGAN to create adversarial fake samples that are classified as real.  To ensure that the images remain realistic, we jointly optimize the classifier loss and GAN loss on the CelebA-HQ dataset:
\begin{equation}
    \mathcal{L} = \min_G \max_D \left[ \mathcal{L}_{\mathrm{GAN}}(G, D) + \mathcal{L}_{\mathrm{real}}(G, C) \right];
\end{equation}
i.e., we optimize both the generator and discriminator with the added constraint that the generator output should be predicted as real by the classifier $C$. Finetuning the generator does not drastically change the generated output (see Supplementary Material Sec.~\ref{apx:finetuning}), but it decreases the classifier's accuracy from 100\% to below 65\% (Fig.~\ref{fig:finetune}). Using a variable threshold (AP) is less sensitive to this adversarial finetuning. We train a second classifier using images from the finetuned generator, which is able to recover in accuracy. We then compute the most predictive image patches for the retrained classifier and cluster them according to semantic category. Compared to the patches captured by the first classifier, this retrained classifier relies less on background patches and more on facial features, suggesting that artifacts in typically solid background patches are easiest for the generator to hide, while artifacts in more textured regions such as hair still remain detectable.


\subsection{Facial manipulation.}

Unlike the fully-generative scenario, facial manipulation methods blend content from two images, hence only a portion of the image is manipulated. Here, we train on each of the four FaceForensics++ datasets~\cite{rossler2019faceforensics++}, and test generalization to the remaining three datasets (Table~\ref{table:faceforensics}). We compare the effect of different receptive fields using truncated models, and investigate which patches are localized.

\setlength{\tabcolsep}{3pt}
\begin{table}[t]
\begin{center}
\caption{Average precision on FaceForensics++~\cite{rossler2019faceforensics++} datasets. Each model is trained on one dataset and evaluated on the remaining datasets.}
\label{table:faceforensics}
\resizebox{0.9\linewidth}{!}{
\begin{tabular}{l cccc cccc}
\toprule
\multicolumn{1}{c}{} & \multicolumn{4}{c}{\textbf{Train on Deepfakes}} & \multicolumn{4}{c}{\textbf{Train on Neural Tex.}}\\
\cmidrule(lr){2-5} \cmidrule(lr){6-9}
\textbf{Model Depth} & \textbf{DF} & \textbf{NT} & \textbf{F2F} & \textbf{FS} & \textbf{DF} & \textbf{NT} & \textbf{F2F} & \textbf{FS}\\
\midrule
Resnet Layer 1 & \gray{98.97} & \textbf{74.99} & \textbf{71.74} & 57.15 & \textbf{70.32} & \gray{86.93} & 65.04 & 52.37 \\
Xception Block 1 & \gray{92.95} & 70.52 & 65.94 & 52.83 & 66.30 & \gray{80.72} & 62.65 & 52.05 \\
Xception Block 2 & \gray{98.04} & 70.28 & 67.48 & 56.04 & 69.61 & \gray{85.75} & 64.27 & 52.70 \\
Xception Block 3 & \gray{99.41} & 67.58 & 63.62 & 57.97 & 67.62 & \gray{85.44} & 60.71 & 52.07 \\
Xception Block 4 & \gray{99.14} & 68.91 & 70.36 & \textbf{58.74} & 73.65 & \gray{90.97} & 60.72 & 52.79 \\
Xception Block 5 & \gray{99.27} & 68.25 & 66.68 & 43.20 & 83.52 & \gray{92.23} & 63.75 & 49.94 \\
\hdashline
\cite{afchar2018mesonet} MesoInception4 & \gray{97.28} & 59.27 & 60.17 & 47.24 & 65.75 & \gray{83.27} & 62.92 & \textbf{54.03} \\
\cite{he2016deep} Resnet-18 & \gray{93.90} & 53.22 & 53.45 & 53.69 & 69.98 & \gray{85.40} & 54.77 & 50.89 \\
\cite{rossler2019faceforensics++} Xception  & \gray{98.60} & 60.15 & 56.84 & 46.12 & 70.07 & \gray{93.61} & 56.79 & 48.55 \\
\cite{wang2019cnn} CNN (p=0.1) & \gray{97.78} & 60.08 & 59.73 & 50.87 & 68.67 & \gray{95.16} & 68.15 & 47.43 \\
\cite{wang2019cnn} CNN (p=0.5) & \gray{98.16} & 54.02 & 56.06 & 55.99 & 66.98 & \gray{95.03} &\textbf{71.50} & 51.93 \\
\bottomrule
\multicolumn{1}{c}{} & \multicolumn{4}{c}{\textbf{Train on Face2Face}} & \multicolumn{4}{c}{\textbf{Train on FaceSwap}}\\
\cmidrule(lr){2-5} \cmidrule(lr){6-9}
\textbf{Model Depth} & \textbf{DF} & \textbf{NT} & \textbf{F2F} & \textbf{FS} & \textbf{DF} & \textbf{NT} & \textbf{F2F} & \textbf{FS} \\
\midrule
Resnet Layer 1 & \textbf{84.39} & 79.72 & \gray{97.66} & 60.53 & 59.49 & 52.56 & 62.00 & \gray{97.13} \\
Xception Block 1 & 77.65 & \textbf{80.88} & \gray{93.84} & 61.62 & 53.14 & 49.24 & 56.89 & \gray{82.89} \\
Xception Block 2 & 84.04 & 79.51 & \gray{97.40} & 63.21 & 58.39 & 51.65 & 61.73 & \gray{92.58} \\
Xception Block 3 & 76.10 & 74.77 & \gray{97.33} & 63.10 & \textbf{61.77} & 53.44 & 61.34 & \gray{96.06} \\
Xception Block 4 & 67.18 & 61.72 & \gray{97.19} & 63.04 & 61.33 & 52.02 & 59.45 & \gray{96.56} \\
Xception Block 5 & 81.25 & 61.91 & \gray{96.45} & 55.15 & 57.14 & 47.39 & 54.68 & \gray{95.57} \\
\hdashline
\cite{afchar2018mesonet} MesoInception4 & 67.53 & 55.17 & \gray{92.27} & 54.06 & 50.64 & 48.87 & 56.15 & \gray{93.81} \\
\cite{he2016deep} Resnet-18 & 55.43 & 52.57 & \gray{93.27} & 53.39 & 61.03 & 51.66 & 52.56 & \gray{91.49} \\
\cite{chollet2017xception} Xception & 66.12 & 56.07 & \gray{97.41} & 53.15 & 53.86 & 50.00 & 56.55 & \gray{96.84} \\
\cite{wang2019cnn} CNN (p=0.1) & 65.76 & 64.81 & \gray{98.40} & 59.48 & 59.19 & \textbf{53.50} &\textbf{ 63.07} & \gray{99.02} \\
\cite{wang2019cnn} CNN (p=0.5) & 65.43 & 60.36 & \gray{97.94} & \textbf{63.52} & 60.19 & 52.11 & 59.81 & \gray{98.25} \\
\bottomrule
\end{tabular}
}
\end{center}
\vspace{-0.32in}
\end{table}
\setlength{\tabcolsep}{1.4pt}

In these experiments, training on Face2Face images yields the best generalization to remaining datasets. On the other hand, generalization to FaceSwap images is the hardest -- training on the other manipulation methods does not generalize well to FaceSwap images, and training on FaceSwap does not generalize well to other manipulation methods. Compared to the full-model baselines, we find that truncated patch classifiers tend to generalize when trained on the Face2Face or Deepfakes domains. Adding augmentations to training \cite{wang2019cnn} can also boost results in some domains. While we do not use mask supervision during training, \cite{li2020face} notes that using this additional supervision signal improves generalization. 


Next, we seek to investigate which patches are identified as predictive using the truncated classifiers in the facial manipulation setting. Unlike the fully generative scenario in which the classifiers tend to focus on the background, these classifiers trained on facial manipulation focus on the face region (without explicit supervision of the face location). In particular, when trained on the Face2Face manipulation method, the classifiers use predominately the mouth region to classify Deepfakes and NeuralTextures manipulation, with eyes or nose as a secondary feature depending on the manipulation method (Fig.~\ref{fig:histograms_F2F}). We show additional visualizations in Supplementary Material Sec.~\ref{apx:faceforensics}.

\begin{figure}[ht!]
\centering
\includegraphics[width=\textwidth]{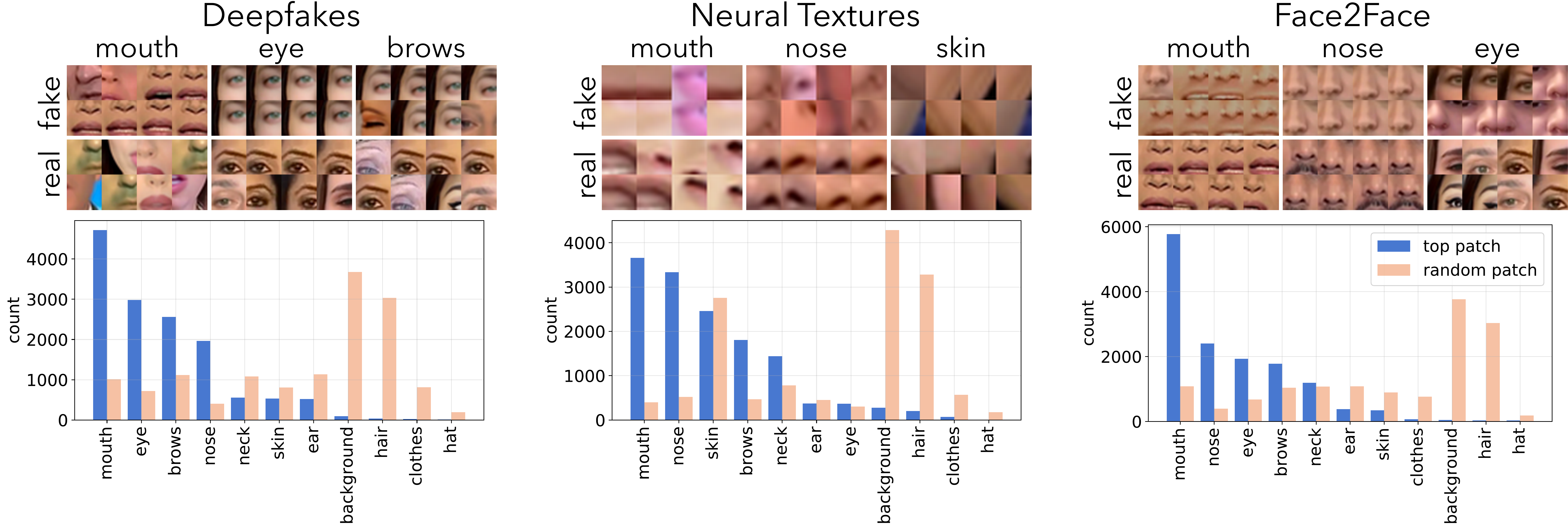} 
\vspace{-0.3in}
\caption{Histograms of the most predictive patches from a classifier trained on Face2Face and un-manipulated images, and tested on the Neural Textures and Deepfakes manipulation methods. Unlike the fully-generative model setup, the classifier in this case localizes patches within the face.}
\label{fig:histograms_F2F}
\end{figure}

\section{Conclusion}
Identifying differences between real and fake images is a constantly evolving problem and is highly sensitive to minor preprocessing details. 
Here, we take the approach of equalizing the preprocessing of the two classes of images to focus on the inherent differences between an image captured from a camera and a doctored image either generated entirely from a deep network, or partially manipulated in facial regions. We investigate using classifiers with limited receptive fields to focus on local artifacts, such as textures in hair, backgrounds, mouths, and eyes, rather than the global semantics of the image. Classifying these small patches allows us to generalize across different model training parameters, generator architectures, and datasets, and provides us with a heatmap to localize the potential areas of manipulation. We show a technique to exaggerate the detectable artifacts of the fake images, and demonstrate that image generators can still be imperfect in certain patches despite finetuning against a given classifier. While progress on detecting fake images inevitably creates a cat-and-mouse problem of using these results to create even better generators, we hope that understanding these detectors and visualizing what they look for can help people anticipate where manipulations may occur in a facial image and better navigate potentially falsified content in today's media.

\paragraph{Acknowledgements.} We thank Antonio Torralba, Jonas Wulff, Jacob Huh, Tongzhou Wang, Harry Yang, and Richard Zhang for helpful discussions. This work was supported by a National Science Foundation Graduate Research Fellowship under Grant No. 1122374 to L.C. and DARPA XAI FA8750-18-C000-4 to D.B.


%
%
\newpage
\bibliographystyle{splncs04}
\bibliography{main}

\begin{thebibliography}{10}
\providecommand{\url}[1]{\texttt{#1}}
\providecommand{\urlprefix}{URL }
\providecommand{\doi}[1]{https://doi.org/#1}

\bibitem{deepfakesgithub}
Deepfakes, \url{https://github.com/deepfakes/faceswap}

\bibitem{afchar2018mesonet}
Afchar, D., Nozick, V., Yamagishi, J., Echizen, I.: Mesonet: a compact facial
  video forgery detection network. In: 2018 IEEE International Workshop on
  Information Forensics and Security (WIFS). pp.~1--7. IEEE (2018)

\bibitem{bau2019seeing}
Bau, D., Zhu, J.Y., Wulff, J., Peebles, W., Strobelt, H., Zhou, B., Torralba,
  A.: Seeing what a gan cannot generate. In: Proceedings of the IEEE
  International Conference on Computer Vision. pp. 4502--4511 (2019)

\bibitem{bayar2016deep}
Bayar, B., Stamm, M.C.: A deep learning approach to universal image
  manipulation detection using a new convolutional layer. In: Proceedings of
  the 4th ACM Workshop on Information Hiding and Multimedia Security. pp. 5--10
  (2016)

\bibitem{carlini2020evading}
Carlini, N., Farid, H.: Evading deepfake-image detectors with white-and
  black-box attacks. In: Proceedings of the IEEE/CVF Conference on Computer
  Vision and Pattern Recognition Workshops. pp. 658--659 (2020)

\bibitem{chollet2017xception}
Chollet, F.: Xception: Deep learning with depthwise separable convolutions. In:
  Proceedings of the IEEE conference on computer vision and pattern
  recognition. pp. 1251--1258 (2017)

\bibitem{cozzolino2017recasting}
Cozzolino, D., Poggi, G., Verdoliva, L.: Recasting residual-based local
  descriptors as convolutional neural networks: an application to image forgery
  detection. In: Proceedings of the 5th ACM Workshop on Information Hiding and
  Multimedia Security. pp. 159--164 (2017)

\bibitem{cozzolino2018forensictransfer}
Cozzolino, D., Thies, J., R{\"o}ssler, A., Riess, C., Nie{\ss}ner, M.,
  Verdoliva, L.: Forensictransfer: Weakly-supervised domain adaptation for
  forgery detection. arXiv preprint arXiv:1812.02510  (2018)

\bibitem{geirhos2018imagenet}
Geirhos, R., Rubisch, P., Michaelis, C., Bethge, M., Wichmann, F.A., Brendel,
  W.: Imagenet-trained cnns are biased towards texture; increasing shape bias
  improves accuracy and robustness. arXiv preprint arXiv:1811.12231  (2018)

\bibitem{goetschalckx2019ganalyze}
Goetschalckx, L., Andonian, A., Oliva, A., Isola, P.: Ganalyze: Toward visual
  definitions of cognitive image properties. In: Proceedings of the IEEE
  International Conference on Computer Vision. pp. 5744--5753 (2019)

\bibitem{gragnaniello2018analysis}
Gragnaniello, D., Marra, F., Poggi, G., Verdoliva, L.: Analysis of adversarial
  attacks against cnn-based image forgery detectors. In: 2018 26th European
  Signal Processing Conference (EUSIPCO). pp. 967--971. IEEE (2018)

\bibitem{hays2007scene}
Hays, J., Efros, A.A.: Scene completion using millions of photographs. ACM
  Transactions on Graphics (TOG)  \textbf{26}(3),  4--es (2007)

\bibitem{he2016deep}
He, K., Zhang, X., Ren, S., Sun, J.: Deep residual learning for image
  recognition. In: Proceedings of the IEEE conference on computer vision and
  pattern recognition. pp. 770--778 (2016)

\bibitem{huh2018fighting}
Huh, M., Liu, A., Owens, A., Efros, A.A.: Fighting fake news: Image splice
  detection via learned self-consistency. In: Proceedings of the European
  Conference on Computer Vision (ECCV). pp. 101--117 (2018)

\bibitem{isola2017image}
Isola, P., Zhu, J.Y., Zhou, T., Efros, A.A.: Image-to-image translation with
  conditional adversarial networks. In: Proceedings of the IEEE conference on
  computer vision and pattern recognition. pp. 1125--1134 (2017)

\bibitem{karras2017progressive}
Karras, T., Aila, T., Laine, S., Lehtinen, J.: Progressive growing of gans for
  improved quality, stability, and variation. arXiv preprint arXiv:1710.10196
  (2017)

\bibitem{karras2019style}
Karras, T., Laine, S., Aila, T.: A style-based generator architecture for
  generative adversarial networks. In: Proceedings of the IEEE Conference on
  Computer Vision and Pattern Recognition. pp. 4401--4410 (2019)

\bibitem{karras2019analyzing}
Karras, T., Laine, S., Aittala, M., Hellsten, J., Lehtinen, J., Aila, T.:
  Analyzing and improving the image quality of stylegan. arXiv preprint
  arXiv:1912.04958  (2019)

\bibitem{kingma2018glow}
Kingma, D.P., Dhariwal, P.: Glow: Generative flow with invertible 1x1
  convolutions. In: Advances in Neural Information Processing Systems. pp.
  10215--10224 (2018)

\bibitem{li2016precomputed}
Li, C., Wand, M.: Precomputed real-time texture synthesis with markovian
  generative adversarial networks. In: European conference on computer vision.
  pp. 702--716. Springer (2016)

\bibitem{li2020face}
Li, L., Bao, J., Zhang, T., Yang, H., Chen, D., Wen, F., Guo, B.: Face x-ray
  for more general face forgery detection. In: Proceedings of the IEEE/CVF
  Conference on Computer Vision and Pattern Recognition. pp. 5001--5010 (2020)

\bibitem{li2018exposing}
Li, Y., Lyu, S.: Exposing deepfake videos by detecting face warping artifacts.
  arXiv preprint arXiv:1811.00656  (2018)

\bibitem{liu2015faceattributes}
Liu, Z., Luo, P., Wang, X., Tang, X.: Deep learning face attributes in the
  wild. In: Proceedings of International Conference on Computer Vision (ICCV)
  (December 2015)

\bibitem{long2015fully}
Long, J., Shelhamer, E., Darrell, T.: Fully convolutional networks for semantic
  segmentation. In: Proceedings of the IEEE conference on computer vision and
  pattern recognition. pp. 3431--3440 (2015)

\bibitem{mayer2019exposing}
Mayer, O., Stamm, M.C.: Exposing fake images with forensic similarity graphs.
  arXiv preprint arXiv:1912.02861  (2019)

\bibitem{mo2018fake}
Mo, H., Chen, B., Luo, W.: Fake faces identification via convolutional neural
  network. In: Proceedings of the 6th ACM Workshop on Information Hiding and
  Multimedia Security. pp. 43--47 (2018)

\bibitem{popescu2004exposing}
Popescu, A.C., Farid, H.: Exposing digital forgeries by detecting duplicated
  image regions. Dept. Comput. Sci., Dartmouth College, Tech. Rep. TR2004-515
  pp. 1--11 (2004)

\bibitem{popescu2005resampling}
Popescu, A.C., Farid, H.: Exposing digital forgeries by detecting traces of
  resampling. IEEE Transactions on signal processing  \textbf{53}(2),  758--767
  (2005)

\bibitem{popescu2005exposing}
Popescu, A.C., Farid, H.: Exposing digital forgeries in color filter array
  interpolated images. IEEE Transactions on Signal Processing  \textbf{53}(10),
   3948--3959 (2005)

\bibitem{rahmouni2017distinguishing}
Rahmouni, N., Nozick, V., Yamagishi, J., Echizen, I.: Distinguishing computer
  graphics from natural images using convolution neural networks. In: 2017 IEEE
  Workshop on Information Forensics and Security (WIFS). pp.~1--6. IEEE (2017)

\bibitem{richardson2018gans}
Richardson, E., Weiss, Y.: On gans and gmms. In: Advances in Neural Information
  Processing Systems. pp. 5847--5858 (2018)

\bibitem{rossler2019faceforensics++}
Rossler, A., Cozzolino, D., Verdoliva, L., Riess, C., Thies, J., Nie{\ss}ner,
  M.: Faceforensics++: Learning to detect manipulated facial images. In:
  Proceedings of the IEEE International Conference on Computer Vision. pp.
  1--11 (2019)

\bibitem{wang2019cnn}
Wang, S.Y., Wang, O., Zhang, R., Owens, A., Efros, A.A.: Cnn-generated images
  are surprisingly easy to spot... for now. arXiv preprint arXiv:1912.11035
  (2019)

\bibitem{xuan2019generalization}
Xuan, X., Peng, B., Wang, W., Dong, J.: On the generalization of gan image
  forensics. In: Chinese Conference on Biometric Recognition. pp. 134--141.
  Springer (2019)

\bibitem{yu2019attributing}
Yu, N., Davis, L.S., Fritz, M.: Attributing fake images to gans: Learning and
  analyzing gan fingerprints. In: Proceedings of the IEEE International
  Conference on Computer Vision. pp. 7556--7566 (2019)

\bibitem{zhang2018unreasonable}
Zhang, R., Isola, P., Efros, A.A., Shechtman, E., Wang, O.: The unreasonable
  effectiveness of deep features as a perceptual metric. In: Proceedings of the
  IEEE conference on computer vision and pattern recognition. pp. 586--595
  (2018)

\bibitem{zhang2019detecting}
Zhang, X., Karaman, S., Chang, S.F.: Detecting and simulating artifacts in gan
  fake images. arXiv preprint arXiv:1907.06515  (2019)

\bibitem{zhou2017two}
Zhou, P., Han, X., Morariu, V.I., Davis, L.S.: Two-stream neural networks for
  tampered face detection. In: 2017 IEEE Conference on Computer Vision and
  Pattern Recognition Workshops (CVPRW). pp. 1831--1839. IEEE (2017)

\end{thebibliography}


\noindent\textbf{\Large{Supplementary Materials}}\\

\noindent In the Supplementary Materials we include additional details on our dataset extraction process for both real and fake images (Sec.~\ref{apx:dataset}--\ref{apx:reprojection}) and details on training the classifiers and visualization (Sec.~\ref{apx:architecture}--\ref{apx:segmentation}). We also include a number of additional experiments and visualizations. Namely, classifiers for generated images are extremely sensitive to subtle differences in preprocessing, and minor differences will allow the classifier to easily learn preprocessing artifacts rather than forensic signals, which gives the appearance of high generalization capability -- we include results that demonstrate this effect in Sec.~\ref{apx:preprocessing}. We also investigate additional training configurations and visualizations in Sec.~\ref{apx:alignment}--\ref{apx:finetuning}.


\section{Supplementary Methods}

\subsection{Dataset Construction}\label{apx:dataset}

\subsubsection{Real Images}

\paragraph{CelebA-HQ:} We prepare the CelebA-HQ images following the dataset preparation pipeline in the PGAN repository.\footnote{\label{fn:pgan}\url{https://github.com/tkarras/progressive_growing_of_gans}}. This saves the images in various resolutions from 4x4px to 1024x1024px in a \texttt{tfrecord} format, which are then subsequently used to train the PGAN generators. For our real CelebA-HQ dataset, we extract the images from the \texttt{tfrecord} format, similar to during PGAN training, at the same resolution of the generator -- i.e., if the fake images are generated from a 128px PGAN, then the real images used for comparison are extracted from the 128px \texttt{tfrecord}. This is to avoid any resizing operations that are different from the subsampling used during GAN training. We partition the images into training and validation images following the CelebA partitions, which yields 24183 images for training and 2993 for validation, and 2824 for testing. All images are resized to 128x128 resolution using Lanczos interpolation and saved in PNG format.

\paragraph{FFHQ:} We download the prepared images in \texttt{tfrecord} format directly from the FFHQ repository\footnote{\url{https://github.com/NVlabs/ffhq-dataset}} and extract them at 1024 resolution for comparison with 1024 resolution generator models. Following the suggested partition in the repository, we reserve the first 60000 images as the training set, the next 5000 for validation and the final 5000 for testing. All images are resized to 128x128 resolution using Lanczos interpolation and saved in PNG format.

\paragraph{CelebA:} We follow the data transformation pipeline used to train the CelebA GMM model.\footnote{\label{fn:gmm}\url{https://github.com/eitanrich/torch-mfa}}, which includes a crop operation to localize the face and bilinear resize to 64 pixel resolution for comparison to the fake images generated by the GMM. We use the testing split of 19962 images for classifier evaluation. All images are resized to 128x128 resolution using Lanczos interpolation and saved in PNG format. The CelebA dataset is a superset of the faces in CelebA-HQ, although at a lower resolution and with a slightly different facial crop. 

\subsubsection{Generated Images}

\paragraph{PGAN:} We download the pretrained 1024px resolution CelebA-HQ generator from the PGAN repository.\footnotemark[\getrefnumber{fn:pgan}] Separately, we train PGANs on the CelebA-HQ dataset to 128px, 256px, and 512px final resolutions. We change the random initialization seed and train 3 additional PGANs to 128px resolution. We also train a PGAN to 1024px resolution on the FFHQ dataset. All images are sampled from the generator, resized to 128x128 resolution using Lanczos interpolation, and saved in PNG format. We sample the same number of images as used in each split of the corresponding real dataset.

\paragraph{SGAN/SGAN2:} We download the pretrained 1024px resolution generators trained on the CelebA-HQ and FFHQ datasets from the StyleGAN repository,\footnote{\url{https://github.com/NVlabs/stylegan}} and the the pretrained 1024px resolution generator trained on the FFHQ dataset from the StyleGAN2 repository.\footnote{\url{https://github.com/NVlabs/stylegan2}} Images are sampled from the generator, resized to 128x128 resolution using Lanczos interpolation, and saved in PNG format. We sample the same number of images as used in each split of the corresponding real dataset.

\paragraph{GMM:} We train the Mixture of Factor Analyzers Gaussian Mixture model based on~\cite{richardson2018gans} using the default training settings provided in the source code.\footnotemark[\getrefnumber{fn:gmm}] Images are sampled from the generator, resized to 128x128 resolution using Lanczos interpolation, and saved in PNG format. We sample the same number of images as used in each split of the corresponding real dataset.

\paragraph{Glow:} We sample images from the 1024px resolution pretrained model in the Glow repository.\footnote{\url{https://github.com/openai/glow}} We perform random manipulation of attributes by selecting an attribute tag and a manipulation amount within the range $\left[-1, 1\right]$ uniformly at random, and apply this to each sample. Images are then resized to 128x128 resolution using Lanczos interpolation and saved in PNG format. We sample the same number of images as used in each split of the corresponding real dataset.

\subsubsection{FaceForensics}~\\
~\\
We download the Deepfakes, Face2Face, FaceSwap, NeuralTextures, and original videos from the FaceForensics++ dataset~\cite{rossler2019faceforensics++}\footnote{\url{https://github.com/ondyari/FaceForensics/}} in compressed format. We use the training and validation splits suggested by the authors. To localize the face, we use the dlib face detector\footnote{\url{https://github.com/davisking/dlib}} to extract eyes, nose, and mouth landmarks, and based on these detected landmarks, align and crop the the frames using the CelebA-HQ alignment approach (the CelebA images are annotated with these landmarks a priori, which are then used to generate the CelebA-HQ images; here, we automate that process). For training images, we extract all frames in the corresponding training videos. For validation and testing images, we extract 100 frames per video to prevent any video from have more influence than the others.

\subsection{Reprojecting Fake Images}\label{apx:reprojection}

Apart from generating fake samples by sampling from a generative model, we can also create hard negative examples by generating fake GAN images that are most similar to a given real target image -- i.e., we project real images to the output manifold of the generative model. To create these images, we use the approach in~\cite{bau2019seeing}, which uses a hybrid encoder and optimization approach. First, given an image $x$, an encoder $E$ is trained layer-wise such that $G(E(x)) \approx x$. This provides a latent code initialization $z = E(x)$ for the second optimization step, which uses an LBGFS optimizer over $z$ to minimize $|x - G(z)|$. We add these reprojected examples to the fake image dataset when training the classifier, but also conduct experiments in which the classifier is trained 1) without reprojected images and 2) only on reprojected images as fake samples in Sec.~\ref{apx:samples_inverse}.

\setlength{\tabcolsep}{3pt}
\begin{table}[h]
\begin{center}
\caption{Model receptive field and parameter count calculations.}
\label{table:architecture}
\vspace{-0.1in}
\resizebox{0.8\linewidth}{!}{
\begin{tabular}{l cr  lr}
\toprule
Truncated Model & RF & \# Params & Full Model & \# Params \\
\cmidrule(lr){1-3} \cmidrule(lr){4-5} 
Resnet Layer 1 & 43 & 0.158 M & \cite{he2016deep} Resnet-18 & 11.178 M \\
Xception Block 1 & 19 & 0.055 M & \cite{afchar2018mesonet} MesoInception4 & 0.029 M\\
Xception Block 2 & 43 & 0.191 M & \cite{chollet2017xception} Xception & 20.811 M \\
Xception Block 3 & 91 & 1.108 M & \cite{wang2019cnn} CNN (p=0.1) & 23.510 M \\
Xception Block 4 & 187 & 2.722 M & \cite{wang2019cnn} CNN (p=0.5) & 23.510 M \\
Xception Block 5 & 263 & 4.336 M &  &  \\
\bottomrule
\end{tabular}}
\end{center}
\vspace{-0.5in}
\end{table}
\setlength{\tabcolsep}{1.4pt}

\subsection{Patch Classifiers Architecture}\label{apx:architecture}

Truncating a classifier reduces the receptive field and number of parameters of the model. Effectively, it increases the ratio of data to model size, as the same model weights are used across all patches of an input image. In Table~\ref{table:architecture}, we provide calculations on the receptive field size and number of parameters for each model used in our experiments. We construct truncated classifiers using Resnet~\cite{he2016deep} and Xception~\cite{chollet2017xception} models as backbones. The Resnet architecture consists of four layers containing residual skip connections -- in our experiments we observe that truncating after the first residual layer often performs best. The Xception architecture consists of 12 blocks with residual connections and separable convolutions, we conduct experiments truncating after the first five blocks. 

\subsection{Additional Training Details}\label{apx:training}

We train all models using the Adam optimizer with default parameters and learning rate (0.001). We use a batch size of 32 images, consisting of 16 real and 16 fake images. After every epoch, we measure raw patch-wise prediction accuracy (without ensembling patch decisions) on validation images corresponding to the training dataset. We stop training when validation accuracy does not improve for a patience parameter of $5*p$ epochs - we use $p=50$ for the Xception Block 1 patch classifier, $p=20$ for the Xception Block 2 patch classifier, and $p=10$ for the deeper models. We use the checkpoint with the highest raw validation accuracy to evaluate on the test data split. For training, we resize all images to native size for each model -- 299 for Xception architectures, 224 for Resnet architectures, and 256 for MesoNet architectures.

When training with reprojected images, we have paired examples of the original image and the reprojected image. When creating batches, we sample both the original and the reprojection within the batch. We use a similar approach for the FaceForensics++ dataset, where we have paired examples of the original and manipulated images. This creates hard negative samples, which we found to improve classification performance. When training with the FaceForensics++ dataset, we do not use information about the manipulated region -- rather, the classifier's objective is simply to predict all patches in the real image as real and all patches in the fake image as fake. Because the background is unlearnable, the classifier's predictions in the background region for the FaceForensics++ datasets remain uncertain. However, \cite{li2020face} notes that using the additional mask location as learning supervision improves classification.

\subsection{Facial Segmentation Model}\label{apx:segmentation}

To visualize the patch-wise model decisions, we use a facial segmentation network and cluster patches by semantic category. Precisely, for each real or fake image, we take the most predictive patch in the image and label it by semantic category. For the segmentation task, we use a BiSeNet pretrained on the CelebAMask-HQ dataset.\footnote{\url{https://github.com/zllrunning/face-parsing.PyTorch}} The network output assigns each input pixel to one of 19 categories: background, skin, left/right brow, left/right eye, eyeglasses, left/right ear, earring, nose, mouth, upper/lower lip, neck, necklace, clothes, hair, and hat. We group together the left/right brow classes into a brow category, left/right eye and eyeglasses into eyes, left/right ear and earring into ears, mouth and upper/lower lip into mouth, and neck and necklace into neck, which yields 11 semantic categories in total. 

Given an image, we use the segmentation network to obtain a semantic class prediction for each pixel. To assign the patch to a cluster, we tally the number of pixels in the patch belonging to each segmentation class, and normalize by the total number of pixels labelled as that class in the image. We assign the patch to the cluster with the highest normalized proportion. This normalization step helps to appropriately weight classes of small features -- e.g. in a patch containing eyebrows, the most common cluster assignment will be skin, but the most common normalized cluster assignment will be eyebrows since the total number of eyebrow pixels in the full image is lower. 


\section{Additional Experiments}

\subsection{Preprocessing}\label{apx:preprocessing}

One critical step of the real vs. fake classification task is to make sure that the classifier is not simply learning differences in the preprocessing of real or fake images, especially since we do not know the exact preprocessing steps taken in the real image datasets. The solution we use is to pass the real images through the data transformation used to train the generator, so that we can construct the real image dataset in a manner as similar as possible to the fake images output from the generator. In our early experiments, we found that a few subtle differences in this preprocessing pipeline will allow the model to learn differences in preprocessing. If all real and fake datasets (training and testing) have this discrepancy, then this will lead to artificially high generalization performance. 

\setlength{\tabcolsep}{3pt}
\begin{table}[hb!]
\begin{center}
\caption{Slight differences in the preprocessing for real or fake images leads to seemingly increases generalization, as the classifier ends up exploiting these differences. We report average precision for a classifier trained on PGAN fake images on the CelebA-HQ dataset and tested on the remaining datasets. AP on the test set corresponding to training images is colored in gray.}
\label{table:preprocessing}
\vspace{-0.2in}
\resizebox{\linewidth}{!}{
\begin{tabular}{lc ccc ccc}
\toprule
&\multicolumn{1}{c}{} & \multicolumn{3}{c}{\textbf{Architectures}} & \multicolumn{3}{c}{\textbf{FFHQ dataset}}\\
\cmidrule(lr){3-5}\cmidrule(lr){6-8}
\textbf{Preprocessing} & \textbf{PGAN} & \textbf{SGAN} & \textbf{GLOW} & \textbf{GMM} & \textbf{PGAN} & \textbf{SGAN} & \textbf{SGAN2} \\
\midrule
identical preprocessing &  \gray{100.00} & 64.80 & 47.06 & 54.69 & 79.20 & 51.15 & 52.37 \\
different interpolation & \gray{99.93} & 99.33 & 99.76 & 85.40 & 99.44 & 99.08 & 99.09 \\
different initial size & \gray{100.00} & 100.00 & 100.00 & 97.81 & 100.00 & 100.00 & 100.00 \\
different image format & \gray{100.00} & 100.00 & 100.00 & 100.00 & 100.00 & 100.00 & 100.00 \\
\bottomrule
\end{tabular}
}
\end{center}
\end{table}
\setlength{\tabcolsep}{1.4pt}

The standard pipeline that we use in the main text for training the fake-image classifier consists of the following steps for real images:
\begin{enumerate}
    \item We pass real images through generator data transform.
    \item All images are resized to the same size (128px) before saving in PNG format.
    \item When loading the image during training, we resize the image to the classifier's native resolution, perform mean centering, and then input to the classifier. 
\end{enumerate}
For the fake images, we replace step 1 with sampling and renormalizing the output from the generator.

We experiment with 3 small variations on the preprocessing pipeline, and apply these variations consistently across the CelebA-HQ, FFHQ, and CelebA real-image datasets.
\begin{itemize}
    \item \underline{interpolation}: Before saving to file, real and fake images undergo different interpolation methods (bilinear vs. lanczos). 
    \item \underline{initial size}: Real and fake images are saved to file in different resolutions.
    \item \underline{image format}: Real and fake images are saved in different image codecs (JPG vs PNG). 
\end{itemize}

\noindent Average precision results using the full Resnet-18 model (generally the weaker model in our generalization experiments) are shown in Table~\ref{table:preprocessing}. When controlling for image format, interpolation, and size differences, the performance of the classifier decreases when tested on different datasets. However, training the same model in which the real dataset is differently processed from the fake dataset leads to the appearance of very high generalization, when in fact the classifier is just learning the differences in preprocessing. These results show that subtle artifacts in preprocessing are in fact easy to learn, and that these classifiers are very sensitive to the types of preprocessing used during training and inference time. Note that these preprocessing artifacts are still learnable even despite a second resizing step which converts real and fake images to the same size with the same interpolation method prior to being input to the classifier.

\subsection{Additional Training Variations}

\subsubsection{Alignment vs Cropping.}\label{apx:alignment} In our main experiments, we train with all patches of aligned facial image and ensemble patch-wise predictions to obtain a binary prediction. We also experimented with random cropping and random resized cropping augmentations during training by resizing the image to 333px resolution, and take a random crop (or random resized crop) of 299 pixels for Xception network training. We compare the classification results on aligned faces, random crops, and random resized crops in Tables~\ref{table:ref1},~\ref{table:randomcrop}, and~\ref{table:randomresizecrop} respectively. The results with added cropping are somewhat mixed -- random cropping boosts Glow generalization, although not as much as training without reprojected images (in Table~\ref{table:samples_only}), and random resized cropping helps slightly in the GMM case, while the FFHQ datasets fare better without cropping augmentation.

\setlength{\tabcolsep}{3pt}
\begin{table}[t!]
\begin{center}

\caption{Average precision on test datasets when trained on CelebA-HQ PGAN images. AP on the test set corresponding to training images is colored in gray.}
\label{table:ref1}
\resizebox{\linewidth}{!}{
\begin{tabular}{lc ccc ccc}
\toprule
&\multicolumn{1}{c}{} & \multicolumn{3}{c}{\textbf{Architectures}} & \multicolumn{3}{c}{\textbf{FFHQ dataset}}\\
\cmidrule(lr){3-5}\cmidrule(lr){6-8}
\textbf{Model} & \textbf{PGAN} & \textbf{SGAN} & \textbf{GLOW} & \textbf{GMM} & \textbf{PGAN} & \textbf{SGAN} & \textbf{SGAN2} \\
\midrule
Xception Block 1 & \gray{100.00} & 98.68 & \textbf{82.39} & 76.21 & 99.68 & 81.35 & 77.40 \\
Xception Block 2 & \gray{100.00} & 99.99 & 46.35 & \textbf{91.38} & \textbf{100.00} & 90.12 & 90.85 \\
Xception Block 3 & \gray{100.00} & \textbf{100.00} & 64.77 & 80.96 & 100.00 & 92.91 & \textbf{91.45} \\
Xception Block 4 & \gray{100.00} & 99.99 & 51.80 & 42.82 & 100.00 & \textbf{95.85} & 90.62 \\
Xception Block 5 & \gray{100.00} & 100.00 & 58.18 & 48.92 & 100.00 & 93.09 & 89.08 \\
\bottomrule
\end{tabular}
}

\vspace{0.2in}

\caption{Average precision on test datasets when trained on CelebA-HQ PGAN images. Random cropping is applied during training.}
\label{table:randomcrop}
\resizebox{\linewidth}{!}{
\begin{tabular}{lc ccc ccc}
\toprule
&\multicolumn{1}{c}{} & \multicolumn{3}{c}{\textbf{Architectures}} & \multicolumn{3}{c}{\textbf{FFHQ dataset}}\\
\cmidrule(lr){3-5}\cmidrule(lr){6-8}
\textbf{Model} & \textbf{PGAN} & \textbf{SGAN} & \textbf{GLOW} & \textbf{GMM} & \textbf{PGAN} & \textbf{SGAN} & \textbf{SGAN2} \\
\midrule
Xception Block 1 & \gray{100.00} & 97.87 & \textbf{89.60} & 84.00 & 99.48 & 79.00 & 76.64 \\
Xception Block 2 & \gray{100.00} & 99.93 & 47.02 & 82.47 & 99.99 & 88.40 & \textbf{89.58} \\
Xception Block 3 & \gray{100.00} &\textbf{ 99.97} & 41.24 & \textbf{86.76} & \textbf{100.00} & 86.69 & 87.25 \\
Xception Block 4 & \gray{100.00} & 99.85 & 56.84 & 66.14 & 100.00 & \textbf{91.60} & 88.36 \\
Xception Block 5 & \gray{100.00} & 99.63 & 54.39 & 84.52 & 99.99 & 89.56 & 88.86 \\
\bottomrule
\end{tabular}
}

\vspace{0.2in}

\caption{Average precision on test datasets when trained on CelebA-HQ PGAN images. Random resized cropping is applied during training.}
\label{table:randomresizecrop}
\resizebox{\linewidth}{!}{
\begin{tabular}{lc ccc ccc}
\toprule
&\multicolumn{1}{c}{} & \multicolumn{3}{c}{\textbf{Architectures}} & \multicolumn{3}{c}{\textbf{FFHQ dataset}}\\
\cmidrule(lr){3-5}\cmidrule(lr){6-8}
\textbf{Model} & \textbf{PGAN} & \textbf{SGAN} & \textbf{GLOW} & \textbf{GMM} & \textbf{PGAN} & \textbf{SGAN} & \textbf{SGAN2} \\
\midrule
Xception Block 1 & \gray{100.00} & 93.22 & 63.99 & 74.13 & 98.81 & 76.83 & 71.04 \\
Xception Block 2 & \gray{100.00} & 99.78 & 46.35 & 56.80 & 99.98 & 85.93 & 85.71 \\
Xception Block 3 & \gray{100.00} & \textbf{99.91} & 50.46 & 56.34 & 99.99 & \textbf{90.10} & \textbf{90.11} \\
Xception Block 4 & \gray{100.00} & 99.80 & 36.94 & 80.88 & \textbf{100.00} & 88.69 & 86.80 \\
Xception Block 5 & \gray{100.00} & 99.79 & \textbf{64.22} & \textbf{92.75} & 99.99 & 89.99 & 89.22 \\
\bottomrule
\end{tabular}
}

\end{center}
\end{table}
\setlength{\tabcolsep}{1.4pt}

\setlength{\tabcolsep}{3pt}
\begin{table}[t!]
\begin{center}
\caption{Average precision on test datasets when trained on Face2Face manipulated images. Images are automatically aligned using facial landmarks. We compare average precision on test images when trained on aligned patches (left four columns) and random crops of patches (right four columns).}
\label{table:randomcrop_faceforensics}
\resizebox{0.9\linewidth}{!}{
\begin{tabular}{l cccc cccc}
\toprule
\multicolumn{1}{c}{} & \multicolumn{4}{c}{\textbf{Face Alignment}} & \multicolumn{4}{c}{\textbf{Random Cropping}}\\
\cmidrule(lr){2-5}\cmidrule(lr){6-9}
\textbf{Model} & \textbf{DF} & \textbf{NT} & \textbf{F2F} & \textbf{FS} & \textbf{DF} & \textbf{NT} & \textbf{F2F} & \textbf{FS} \\
\midrule
Xception Block 1 & 77.65 & \textbf{80.88} & \gray{93.84} & 61.62 & 76.46 & 79.11 & \gray{92.44} & 60.08 \\
Xception Block 2 &\textbf{ 84.04} & 79.51 & \gray{97.40} & \textbf{63.21} & \textbf{83.19} & \textbf{80.61} & \gray{97.01} & \textbf{63.59} \\
Xception Block 3 & 76.10 & 74.77 & \gray{97.33} & 63.10 & 76.15 & 75.23 & \gray{98.00} & 63.17 \\
Xception Block 4 & 67.18 & 61.72 & \gray{97.19} & 63.04 & 71.91 & 68.01 & \gray{98.60} & 62.01 \\
Xception Block 5 & 81.25 & 61.91 & \gray{96.45} & 55.15 & 76.44 & 62.78 & \gray{96.10} & 52.63 \\
\bottomrule
\end{tabular}
}
\end{center}
\end{table}
\setlength{\tabcolsep}{1.4pt}

A possible explanation for this behavior is that faces are naturally structured and can be aligned via facial landmarks -- CelebA-HQ and FFHQ datasets are already aligned, whereas we automatically align the FaceForensics dataset using a facial landmark detector (see Table~\ref{table:randomcrop_faceforensics} for a comparison). Secondly, truncated models can tolerate minor shifts, as the same model weights are applied over local patches in a sliding fashion over the image. 

\newpage
\subsection{Receptive field vs. number of parameters}\label{apx:rf} 

There are two main differences between a shallow truncated model and a deeper one -- receptive field and number of parameters. Hypothetically, both factors could reduce overfitting and improve generalization. In Table~\ref{table:rfmodel} we seek disentangle these two components when evaluating generalization on test datasets. We take the Xception Block 2 and Xception Block 4 truncated models, and additionally train an Extended Block 2 model by adding 2 additional Xception blocks to the Block 2 model, modified with 1x1 convolutions. This increases the number of parameters of the Block 2 truncated model without increasing the receptive field. However, despite this increase in parameters, we do not see large increases in average precision on the test datasets, suggesting that perhaps the receptive field size contributes more to generalization on unseen faces at test time.

\setlength{\tabcolsep}{3pt}
\begin{table}[t!]
\begin{center}
\caption{Comparing the effect on model size and receptive field on test datasets. The Extended block2 model adds two additional Xception blocks modified with 1x1 convolutions to increase parameter count without increasing receptive field. AP on the test set corresponding to training images is colored in gray.}
\vspace{-0.1in}
\label{table:rfmodel}
\resizebox{\linewidth}{!}{
\begin{tabular}{lc ccc ccc}
\toprule
&\multicolumn{1}{c}{} & \multicolumn{3}{c}{\textbf{Architectures}} & \multicolumn{3}{c}{\textbf{FFHQ dataset}}\\
\cmidrule(lr){3-5}\cmidrule(lr){6-8}
\textbf{Model} & \textbf{PGAN} & \textbf{SGAN} & \textbf{GLOW} & \textbf{GMM} & \textbf{PGAN} & \textbf{SGAN} & \textbf{SGAN2} \\
\midrule

Xception Block 2 & \gray{100.00} & 99.99 & 46.35 & 91.38 & 100.00 & 90.12 & \textbf{90.85} \\
Extended Block 2 & \gray{100.00} & 99.99 & 39.23 & \textbf{91.54} & 100.00 & 87.82 & 89.43 \\
Xception Block 4 & \gray{100.00} & 99.99 & \textbf{51.80} & 42.82 & 100.00 & \textbf{95.85} & 90.62 \\

\bottomrule
\end{tabular}
}
\vspace{-0.2in}
\end{center}
\end{table}
\setlength{\tabcolsep}{1.4pt}

\subsection{Training with reprojected image samples.}\label{apx:samples_inverse} 

For detecting fake images created from generative models, we find that adding reprojected fake images -- i.e., the GAN-generated image most similar a given real image, helps to improve performance on the test datasets in most cases. Here, we compare those results (Table~\ref{table:reference2}) to models trained only on random samples from the fake image generators  (Table~\ref{table:samples_only}) and models trained only on reprojected fake images (Table~\ref{table:inverses_only}). One exception where adding the reprojected fake images harms classification, compared to training only with random samples, in the case of the Glow fake images where there is a difference in AP of over 10\%. On the other hand, when training only with reprojected images there is a domain shift between train and test data, as the model is evaluated on random samples from the generator which are never seen during training. Hence, the test AP is lower in most cases, except for the GMM model where it is similar.

\begin{figure}[ht!]
\centering
\includegraphics[width=\textwidth]{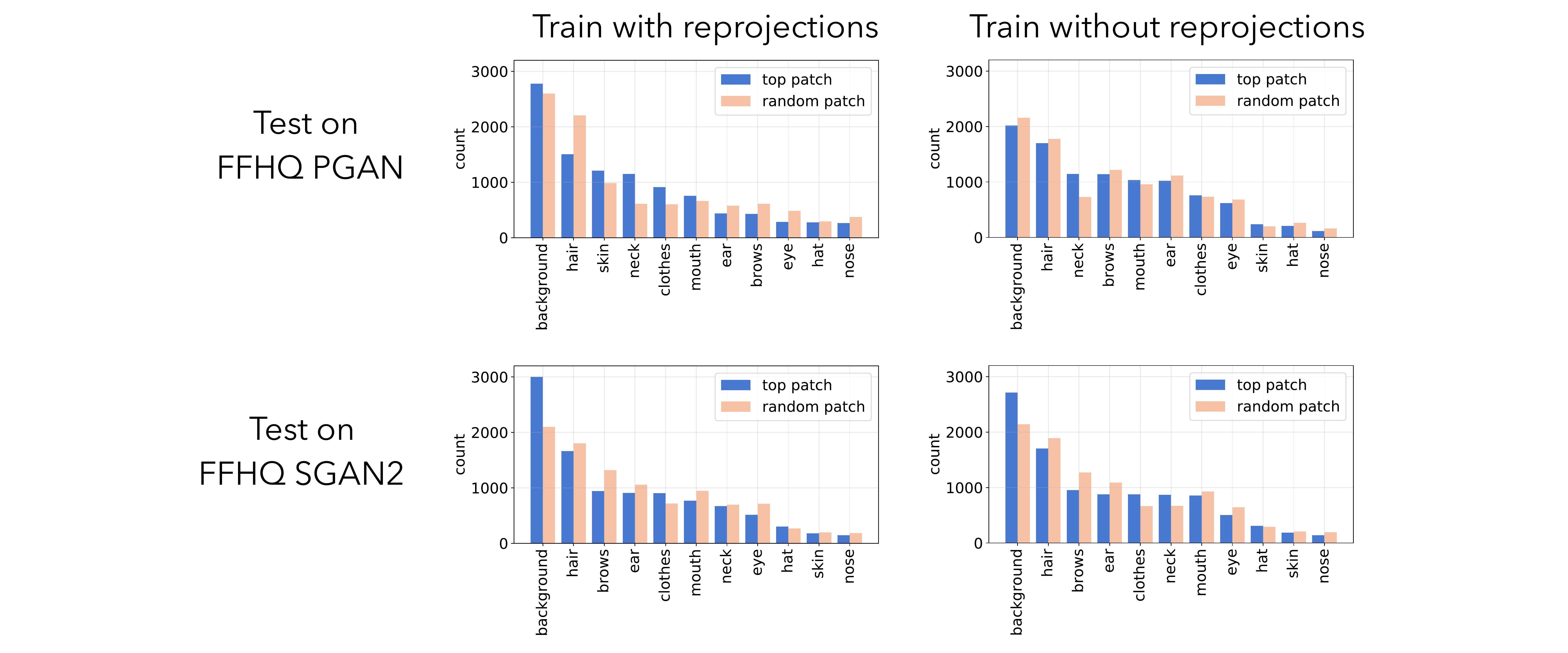}
\caption{Top patches categorized by segmentation category for classifiers trained with and with reprojected fake images. Adding reprojections to the training set causes the classifier to place greater emphasis on background patches.
\label{fig:samples_histograms}}
\end{figure}

How does training with reprojected images affect the patches that the classifier uses for classification? For this experiment, we take the same classifiers as before, trained on CelebA-HQ PGAN images. In Fig.~\ref{fig:samples_histograms} we show patches grouped by semantic category when the classifier is tested on various FFHQ generators. Training with reprojected images causes the classifier place greater emphasis on background patches, compared to training without reprojections, suggesting that adding the reprojection better allows the classifier to learn artifacts in the background portion of the image. 

\setlength{\tabcolsep}{3pt}
\begin{table}[ht!]
\begin{center}
\caption{Average precision on datasets when trained on CelebaHQ PGAN images and reprojected images as the fake image dataset. AP on the test set corresponding to training images is colored in gray.}
\label{table:reference2}
\resizebox{\linewidth}{!}{
\begin{tabular}{lc ccc ccc}
\toprule
&\multicolumn{1}{c}{} & \multicolumn{3}{c}{\textbf{Architectures}} & \multicolumn{3}{c}{\textbf{FFHQ dataset}}\\
\cmidrule(lr){3-5}\cmidrule(lr){6-8}
\textbf{Model} & \textbf{PGAN} & \textbf{SGAN} & \textbf{GLOW} & \textbf{GMM} & \textbf{PGAN} & \textbf{SGAN} & \textbf{SGAN2} \\
\midrule
Xception Block 1 & \gray{100.00} & 98.68 & \textbf{82.39} & 76.21 & 99.68 & 81.35 & 77.40 \\
Xception Block 2 & \gray{100.00} & 99.99 & 46.35 & \textbf{91.38} & \textbf{100.00} & 90.12 & 90.85 \\
Xception Block 3 & \gray{100.00} & \textbf{100.00} & 64.77 & 80.96 & 100.00 & 92.91 & \textbf{91.45} \\
Xception Block 4 & \gray{100.00} & 99.99 & 51.80 & 42.82 & 100.00 & \textbf{95.8}5 & 90.62 \\
Xception Block 5 & \gray{100.00} & 100.00 & 58.18 & 48.92 & 100.00 & 93.09 & 89.08 \\
\bottomrule
\end{tabular}
}

\vspace{0.2in}

\caption{Average precision on datasets when trained on only CelebA-HQ PGAN samples as the fake image dataset.}
\label{table:samples_only}
\resizebox{\linewidth}{!}{
\begin{tabular}{lc ccc ccc}
\toprule
&\multicolumn{1}{c}{} & \multicolumn{3}{c}{\textbf{Architectures}} & \multicolumn{3}{c}{\textbf{FFHQ dataset}}\\
\cmidrule(lr){3-5}\cmidrule(lr){6-8}
\textbf{Model} & \textbf{PGAN} & \textbf{SGAN} & \textbf{GLOW} & \textbf{GMM} & \textbf{PGAN} & \textbf{SGAN} & \textbf{SGAN2} \\
\midrule
Xception Block 1 & \gray{100.00} & 93.78 & \textbf{95.48} & \textbf{78.91} & 89.29 & 67.84 & 66.74 \\
Xception Block 2 & \gray{100.00} & 99.90 & 67.49 & 77.34 & 99.89 & 84.27 & 84.56 \\
Xception Block 3 & \gray{100.00} & \textbf{99.92} & 74.98 & 71.29 & \textbf{99.97} & \textbf{88.49} & \textbf{87.78} \\
Xception Block 4 & \gray{100.00} & 98.81 & 66.79 & 68.06 & 99.79 & 84.67 & 79.50 \\
Xception Block 5 & \gray{100.00} & 95.25 & 60.44 & 68.47 & 98.95 & 71.75 & 70.83 \\
\bottomrule
\end{tabular}
}

\vspace{0.2in}

\caption{Average precision on datasets when trained on only images reprojected via PGAN as the fake image dataset.}
\label{table:inverses_only}
\resizebox{\linewidth}{!}{
\begin{tabular}{lc ccc ccc}
\toprule
&\multicolumn{1}{c}{} & \multicolumn{3}{c}{\textbf{Architectures}} & \multicolumn{3}{c}{\textbf{FFHQ dataset}}\\
\cmidrule(lr){3-5}\cmidrule(lr){6-8}
\textbf{Model} & \textbf{PGAN} & \textbf{SGAN} & \textbf{GLOW} & \textbf{GMM} & \textbf{PGAN} & \textbf{SGAN} & \textbf{SGAN2} \\
\midrule
Xception Block 1 & \gray{97.74} & 90.57 & 31.30 & 77.49 & 99.60 & 70.92 & 71.90 \\
Xception Block 2 & \gray{99.98} & \textbf{99.34} & 39.91 & \textbf{92.03} & \textbf{99.97} & \textbf{84.00} & \textbf{84.27} \\
Xception Block 3 & \gray{99.86} & 99.23 & 45.53 & 89.89 & 99.95 & 79.90 & 78.57 \\
Xception Block 4 & \gray{99.02} & 97.13 & 48.06 & 43.24 & 99.21 & 66.00 & 66.60 \\
Xception Block 5 & \gray{87.30} & 79.86 & 48.94 & 50.47 & 96.25 & 53.31 & 56.65 \\
\bottomrule
\end{tabular}
}

\end{center}
\end{table}
\setlength{\tabcolsep}{1.4pt}

\newpage
\subsection{Investigating biases in the classifiers}

To investigate biases in the fake-image classifier, we take the pre-trained detector on male and female faces from~\cite{karras2019style} and compute average precision conditioned on the male or female classes predicted by the detector. Classifiers are on trained on CelebA-HQ faces and PGAN samples. When there is no domain gap between training and test time, the classifier can solve the task perfectly for both male and female categories. Next we test on two datasets where a domain gap exists -- CelebA-HQ faces generated using the Glow model and FFHQ faces generated using the SGAN2 model. In these more difficult cases, the fake-image classifier obtains slightly higher AP on faces categorized as male by the pre-trained detector.

\setlength{\tabcolsep}{3pt}
\begin{table}[t!]
\begin{center}
\caption{We take a pre-trained classifier on male and female faces and calculate AP on the test images to investigate biases in the fake-image classifier.}
\vspace{-0.1in}
\label{table:biases}
\resizebox{\linewidth}{!}{
\begin{tabular}{l cc cc cc}
\toprule
\textbf{Test Set} & \textbf{AP Overall} & \textbf{\# Total} & \textbf{AP male} & \textbf{\# Male} & \textbf{AP female}  & \textbf{\# Female}\\
\midrule
CelebAHQ PGAN & 100.0 & 5986 & 100.0 & 2024 & 100.0 & 3962 \\
CelebAHQ Glow & 94.9 & 5986 & 96.6 & 2034 & 93.9 & 3952 \\
FFHQ SGAN2 & 91.7 & 10000 & 93.3 & 4480 & 90.4 & 5520 \\
\bottomrule
\end{tabular}
}
\vspace{-0.2in}
\end{center}
\end{table}
\setlength{\tabcolsep}{1.4pt}

\subsection{Additional FaceForensics Visualizations}\label{apx:faceforensics}

\begin{figure}[ht!]
\centering
\includegraphics[width=\textwidth]{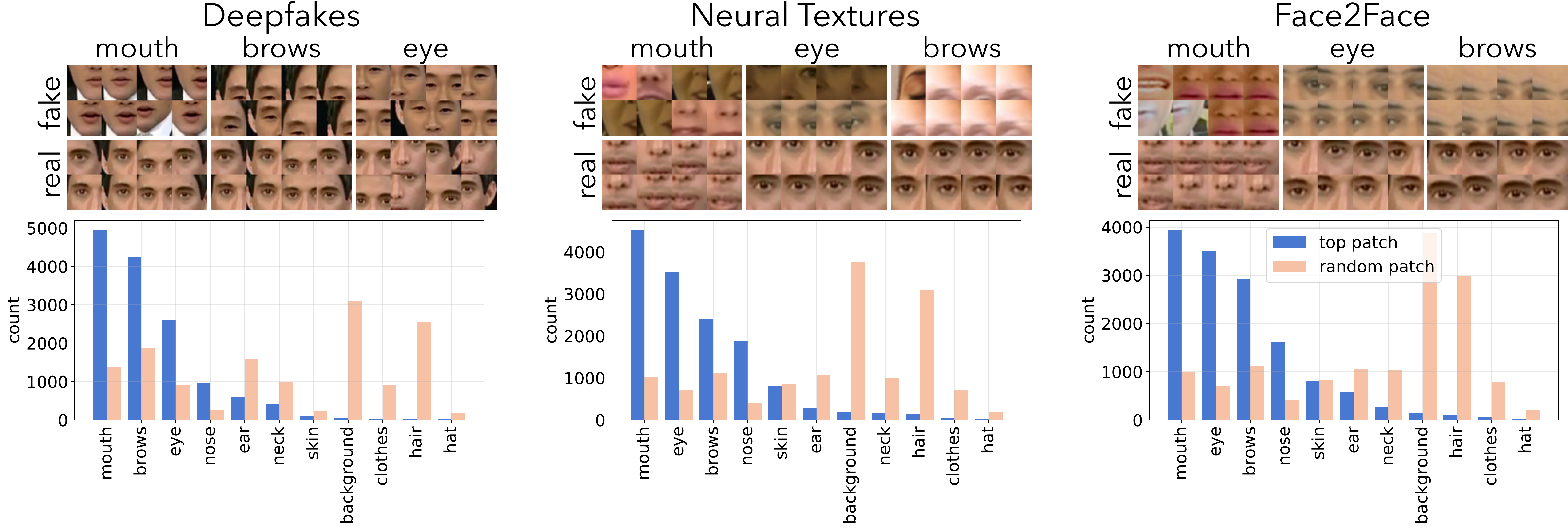} 
\caption{Histograms of the most predictive patches from a classifier trained on Deepfakes and un-manipulated images, and tested on the Neural Textures and Face2Face manipulation methods
\label{fig:histograms_DF}}
\end{figure}

\begin{figure}[ht!]
\centering
\includegraphics[width=\textwidth]{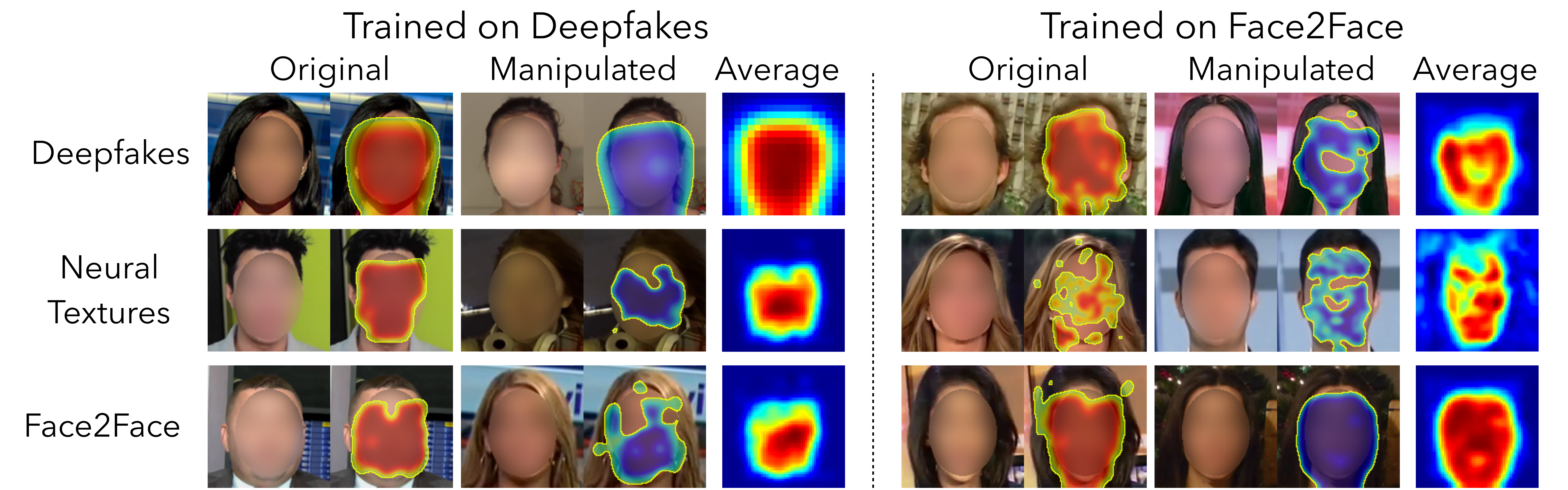}
\caption{Heatmaps showing examples of the patch-wise prediction output of a classifier on one FaceForensics method and tested on other methods. We also show the average heatmap over the 100 most predictive real and manipulated images for each dataset.
\label{fig:heatmaps_faceforensics}}
\end{figure}

In the main text we show patch-wise visualizations and statistics for training on unmanipulated images and Face2Face images and testing on Neural Textures and Deepfakes images. Here we show similar visualization when trained on Deepfakes images in Fig.~\ref{fig:histograms_DF}. In addition, we show examples of local classifier predictions and heatmaps of the top 100 most predictive images in Fig~\ref{fig:heatmaps_faceforensics}. While the heatmap of test images corresponding to the training set capture the general face area, heatmaps of images corresponding to different manipulation methods highlight more local features, such as lower face or eye regions. 

\newpage
\subsection{Example Images after Finetuning}\label{apx:finetuning}
\begin{figure}[!ht]
\centering
\includegraphics[width=0.45\linewidth]{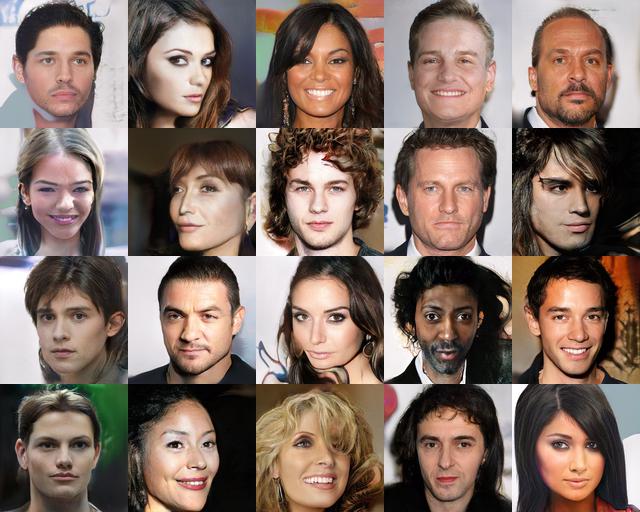}
\hspace{0.1in}
\includegraphics[width=0.45\linewidth]{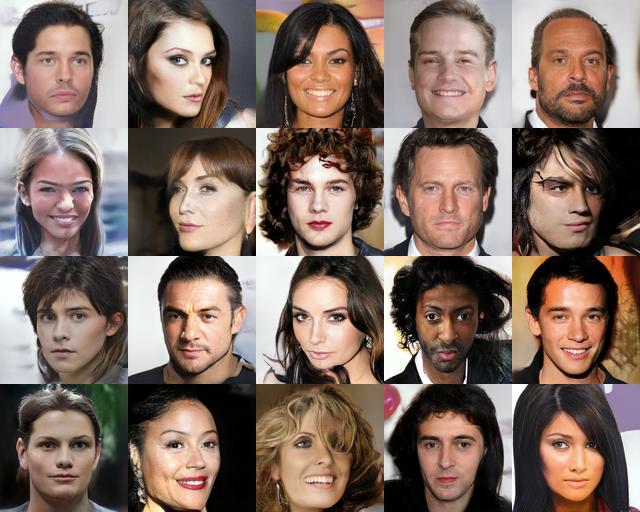}
\caption{Samples from PGAN generator before (left) and after (right) finetuning to evade a fakeness classifier.}
\label{fig:finetune-samples}
\end{figure}

In the main text, we finetune a face PGAN generator to evade classification by a fakeness detector, which drops detection accuracy to from 100\% to below 65\%. Fig.~\ref{fig:finetune-samples} shows random samples from the generator before and after finetuning. The samples remain visually similar despite finetuning but are misclassified by the detector. We further show in the main text that a secondary classifier trained on these finetuned images can recover in classification accuracy, which suggests that finetuning does not completely remove the detectable artifacts and the finetuned images are still distinguishable from real faces.

\newpage

\end{document}